\documentclass[11pt]{article}
\usepackage[T1]{fontenc}
\usepackage[utf8]{inputenc}
\usepackage[margin=1in]{geometry}
\usepackage{amsmath,amssymb}
\usepackage{graphicx}
\usepackage{booktabs}
\usepackage{longtable}
\usepackage{array}
\usepackage{pdflscape}
\usepackage{caption}
\usepackage{float}
\usepackage[hidelinks]{hyperref}
\makeatletter
\@ifundefined{Gin@key@alt}{\define@key{Gin}{alt}{}}{}
\makeatother

\title{Candidate supply and answer selection shape the value of LLM judging in multi-agent systems}
\author{Jia-Hao Ji\textsuperscript{1\#}, Sijie Li\textsuperscript{2,4\#}, Jiabei Cheng\textsuperscript{2,3\#}, Zixi She\textsuperscript{2}, Jin-Tai Yu\textsuperscript{1*}, Zhiyuan Yuan\textsuperscript{2,4*}}
\date{}

\begin{document}
\maketitle
\begin{center}
\small
\textsuperscript{1}Department of Neurology and National Center for Neurological Diseases, Huashan Hospital, State Key Laboratory of Medical Neurobiology and Ministry of Education Frontiers Center for Brain Science, Shanghai Medical College, Fudan University, Shanghai 200040, China.\par
\textsuperscript{2}Institute of Science and Technology for Brain-Inspired Intelligence, MOE Key Laboratory of Computational Neuroscience and Brain-Inspired Intelligence, MOE Frontiers Center for Brain Science, Fudan University, Shanghai 200433, China.\par
\textsuperscript{3}School of Automation and Intelligent Sensing, Shanghai Jiao Tong University, Shanghai 200433, China.\par
\textsuperscript{4}Shanghai Academy of Artificial Intelligence for Science, Shanghai 200003, China.\par
\medskip
\textsuperscript{\#}These authors contributed equally.\\\textsuperscript{*}Corresponding authors: jintai\_yu@fudan.edu.cn (J.-T.Y.); zhiyuan@fudan.edu.cn (Z.Y.)
\end{center}
\begin{abstract}
Multi-agent systems (MAS) sometimes already have the potential to answer correctly, but still report a wrong answer. Explaining this outcome is difficult because generation, communication and final answer-selection rules usually change simultaneously. We conceptualize multi-agent reasoning as an evolutionary pipeline of candidate generation, peer communication and terminal selection, wherein consensus without quality control can exhibit patterns of memetic drift. We study two questions: (1) when an LLM judge provides effective selection pressure by supplying a signal of answer correctness for candidates generated in a multi-agent system, and (2) when using that signal improves the reported answer. To map judge reliability, we analysed 15,336 questions from MMLU-Pro, GPQA, MedXpertQA and MuSR, with Humanity's Last Exam analysed separately. To test these rules, we replayed 81,390 fixed candidate pools drawn from 16,278 questions across five benchmarks. We report three findings. (1) A correct answer is often already present among the generated candidates, but the system can still converge on and report a wrong answer. (2) Judge reliability is not a fixed trait of the model, but varies with the task, the generator and how rare the correct answer is. (3) Combining answer frequency with the judge's evaluation changed only the final answer-selection rule and raised accuracy from 63.82\% to 70.82--70.95\%, primarily by rescuing correct answers that were outnumbered by popular errors. In the systems studied here, the value of generating more candidates depends on whether those extra samples make correct answers present, frequent or recognisable. By isolating generation, recognition and selection, these findings establish a diagnostic basis for designing multi-agent architectures that protect generated correct answers from being lost.
\end{abstract}
\section{Introduction}
\label{introduction}
Multi-agent systems (MAS) are increasingly used to reason, plan, write software and coordinate complex tasks\cite{ref1,ref2,ref3,ref4,ref5,ref6}. Du \emph{et al.} let model instances exchange arguments\cite{ref7}. Chen \emph{et al.} brought diverse models into a confidence-weighted round table\cite{ref8}. Hong \emph{et al.} assigned specialised software-development roles\cite{ref9}, while other systems routed information through dynamic or searched workflows\cite{ref10,ref11,ref12,ref13}. Most comparisons change candidate generation, communication, topology and answer selection at the same time, so a higher score does not show which operation helped\cite{ref14,ref15,ref16}. This inability to attribute system-level gains has become a central transparency problem for multi-agent AI\cite{ref17}. The resulting opacity conceals a simple but consequential paradox. A group of agents can collectively generate a correct answer while the system still reports a wrong one.

A correct answer can be generated and still fail to become the reported output. Apparent gains may come from additional sampling, interaction among agents, or answer selection (the rule used to turn their responses into one reported answer). Choi \emph{et al.} found that majority voting accounted for most of the gains typically attributed to multi-agent debate\cite{ref18}, consistent with findings from Wang \emph{et al.} and Li \emph{et al.} that independent sampling and voting can improve reasoning without communication\cite{ref19,ref20}. This does not make communication irrelevant. Smit \emph{et al.} found that its benefits were not reliable across settings, although tuned protocols could help\cite{ref21}, and Kim \emph{et al.} showed that its value varies with the task, model capability and system architecture under controlled token budgets\cite{ref14}. Nor does adding more agents or independently generated answers ensure that the group reports the best available answer. When agents repeat the same plausible error, voting can suppress a less frequent correct alternative\cite{ref22}. A signal that helps distinguish correct from incorrect answers can instead give that outnumbered answer a path to the reported output. Recent work has pursued this idea through structured reasoning-tree auditing\cite{ref23} and hidden-state selection\cite{ref24}.

To understand why a correct answer can be lost after being generated, we first follow what happens to information through the system. We formalize multi-agent reasoning as a modular information pipeline with three functional phases. (1) During candidate generation, agents independently produce diverse answer candidates and supporting rationales. (2) During information evolution, peer interaction allows agents to repeat, adopt, revise or combine statements. (3) During terminal selection, the system aggregates or filters candidates to produce a final output\cite{ref7,ref8,ref25,ref26,ref27} (\hyperref[terminology]{see Methods}). When agents exchange outputs without a quality filter, stochastic variation can give one rationale an early advantage, a pattern analogous to \emph{memetic drift}\cite{ref28}. Repeated peer adoption can then amplify that advantage into a majority-biased cascade\cite{ref29,ref30,ref31,ref32}. Memetic drift and majority-biased cascades can therefore coexist and jointly create a generation--retention bottleneck in which a correct answer remains available but a popular error gains collective support. To counteract this bottleneck, the system may benefit from selection pressure that evaluates candidate validity independently of popularity. An external verifier or LLM judge can exert this selection pressure by evaluating the structural coherence of individual rationales. In practical settings, quality may be defined by human preference or by whether an action achieves its intended outcome. Labelled reasoning benchmarks offer a clearer case because each question has a predefined correct answer. However, comparisons among multi-agent systems often vary several mechanisms together\cite{ref14,ref15,ref16}, making it difficult to tell whether a system succeeds by distinguishing correct answers or merely by amplifying popularity. We therefore separate whether a judge can recognise correct answers from how an answer-selection rule uses that signal. We ask when an LLM judge evaluates candidates reliably and when an answer-selection rule can use those evaluations to help a correct answer overcome a popular error.

When direct code execution\cite{ref33,ref34} or domain-specific verifiers\cite{ref35,ref36,ref37} are unavailable, LLM judges provide a scalable alternative for exerting selection pressure on candidate answers and rationales\cite{ref38,ref39,ref40,ref41,ref42,ref43}. Their judgments are nevertheless susceptible to systematic biases, including position bias\cite{ref44}, length bias\cite{ref45} and self-preference\cite{ref46}. Judge reliability therefore depends on the candidates being evaluated and should be assessed in the specific generator, task and correct-answer availability regime in which the judge will be used.

These distinctions yield three empirical predictions. Prediction 1: multi-agent review can report a wrong answer even when a correct candidate has already been generated, indicating a retention bottleneck after generation. Prediction 2: judge reliability (the efficacy of selection pressure) should vary with correct-answer availability and with the task, candidate generator and inference setting. Prediction 3: using an informative judge signal for terminal selection should help most when correct candidates are available but outnumbered, and less when majority voting is already reliable. We test these predictions separately by tracking whether generated correct candidates reach the reported answer, measuring correct-versus-wrong ranking performance and comparing answer-selection rules on identical stored candidate pools.

We test these predictions in four stages. First, we analyse multi-agent medical reasoning protocols. Second, we map judge reliability across 15,336 questions from MMLU-Pro\cite{ref47}, GPQA\cite{ref48}, MedXpertQA\cite{ref49} and MuSR\cite{ref50}, with Humanity's Last Exam analysed separately\cite{ref51}. Third, we test alternative answer-selection rules on 81,390 fixed candidate pools. Finally, we reconstruct token-based cost scaling.

These stages supported the three predictions within the tested systems. In the medical experiments, review protocols improved accuracy only slightly despite using more tokens than majority voting. Correct answers were nevertheless present on many questions for which the system reported a wrong answer, locating the loss after generation. Across the stored candidate banks, ranking reliability varied with the task, generator, inference setting and correct-answer availability. On the same fixed pools, combining answer frequency with complete candidate rankings reached 70.82--70.95\% accuracy, compared with 63.82\% for majority vote. Together, these experiments show when a correct candidate that has already been generated can be recognised and preserved in the system's final output.

\section{Results}
\label{results-heading}
\subsection{Multi-agent review does not reliably retain generated correct answers}
\begin{figure}[H]
\centering
\phantomsection\label{fig1}
\includegraphics[alt={Fig. 1 | Multi-agent systems can generate correct answers yet lose them downstream when communication and aggregation favour popular errors over less-frequent correct alternatives},width=\linewidth]{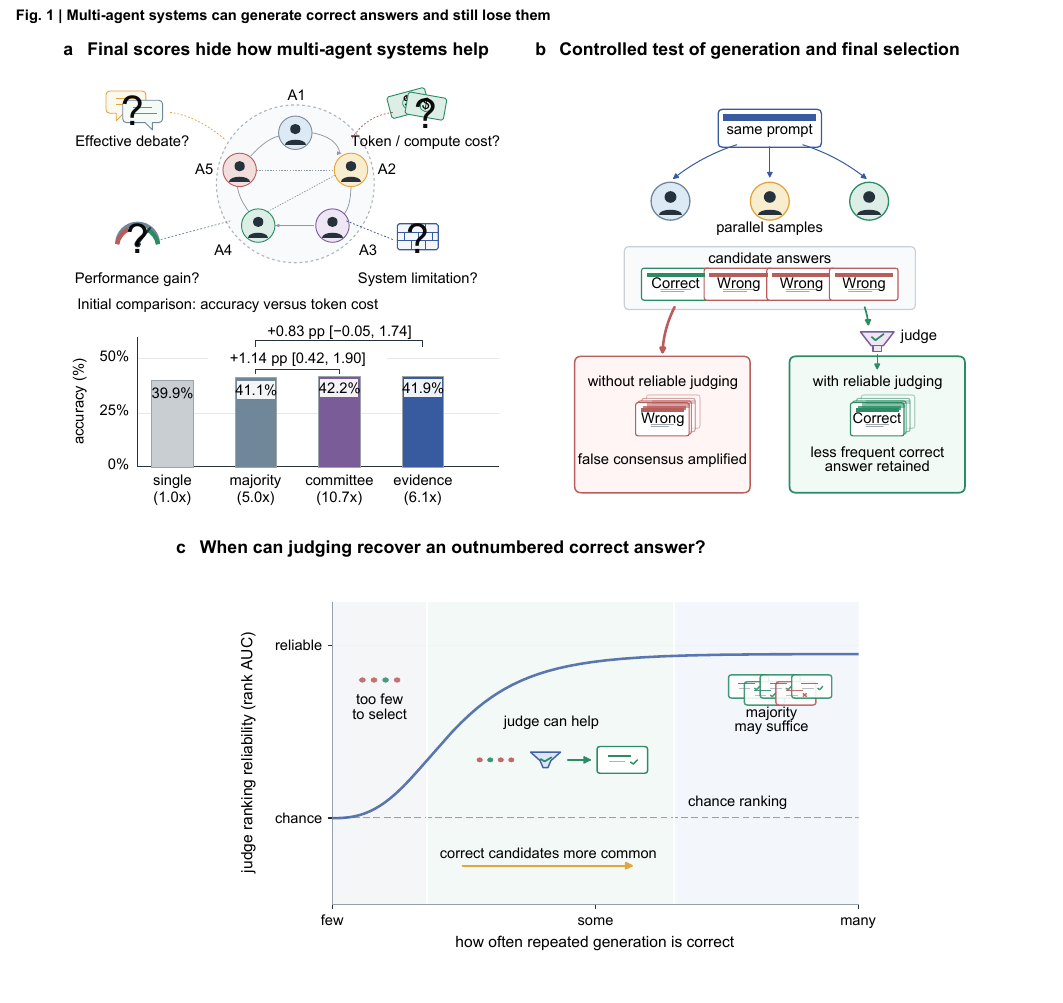}
\caption*{\textbf{Fig. 1 | Multi-agent systems can generate correct answers yet lose them downstream when communication and aggregation favour popular errors over less-frequent correct alternatives}}
\end{figure}

Multi-agent communication protocols incur substantial computational overhead while failing to reliably retain generated correct answers in the reported output (Prediction 1). To isolate this generation--retention bottleneck, we analysed 7,350 case runs per protocol across 2,450 medical reasoning questions from MedXpertQA under four controlled answer-generation and interaction protocols: single-answer generation (1.00\ensuremath{\times} cost baseline), five-answer majority voting without communication, an open committee with source-labelled peer review, and an anonymous evidence board pooling unlinked arguments (\hyperref[fig1]{Fig. 1a}; \hyperref[motivation]{see Methods, Initial MAS architecture and discussion experiments}).

The communication protocols used more tokens than majority voting but yielded little accuracy gain. A single response achieved 39.88\% accuracy at 1.00\ensuremath{\times} relative token usage, whereas majority voting reached 41.06\% accuracy at 4.96\ensuremath{\times} token cost. The open committee reached 42.20\% accuracy at 10.73\ensuremath{\times} token cost, 1.14 percentage points higher than majority voting. The corresponding 95\% confidence interval (CI) was 0.42 to 1.90 percentage points. The evidence board reached 41.89\% accuracy at 6.07\ensuremath{\times} token cost, 0.83 percentage points higher than majority voting (95\% CI, \ensuremath{-}0.05--1.74; \hyperref[fig1]{Fig. 1a}). The open committee therefore used more than twice as many tokens as majority voting for a small accuracy gain. The evidence board also used extra computation, but its confidence interval included zero, so the data were compatible with no difference from majority voting.

Available correct candidates were often not preserved in the reported output, leaving a 13.13--14.38 percentage point ``generation--retention gap''. To isolate candidate generation failure from downstream loss after generation, we calculated an oracle upper bound by scoring a question as correct whenever at least one initial response contained the correct answer. Across multi-answer protocols, this oracle ceiling exceeded actual system accuracy by 13.13--14.38 percentage points (\hyperref[fig1]{Fig. 1a} and \hyperref[ed_fig1]{Extended Data Fig. 1d}). Crucially, on questions where the most common initial answer was incorrect and an outnumbered correct candidate was also present, the open committee selected the most frequent incorrect answer in 73.25\% of cases, and the evidence board did so in 59.61\% of cases (\hyperref[ed_fig1]{Extended Data Fig. 1d}). This is a retention failure: the system can generate a correct minority answer but still report the popular error.

Longer multi-round discussion did not resolve this bottleneck. In an independent sweep over four five-agent communication topologies (star, complete, ring and hierarchical), 90.48--95.37\% of runs ended after one round under the standard convergence rules (\hyperref[ed_fig1]{Extended Data Fig. 1b}). Forcing every system to complete six full rounds increased token use to 33\ensuremath{\times}--49\ensuremath{\times} without consistent accuracy gains across topologies (\hyperref[ed_fig1]{Extended Data Fig. 1a}). This rapid convergence and the absence of benefit from longer discussion were consistent with the persistence of the oracle gap after a correct answer had been generated. These results support Prediction 1: across the tested protocols, correct initial answers were available in cases where the reported output was wrong (\hyperref[fig1]{Fig. 1b} and \hyperref[ed_fig1]{Extended Data Fig. 1d}).

Together, these multi-agent interaction experiments show a repeated failure mode: unconstrained communication can discard minority correct answers as popular errors gain collective support. In our framework, memetic drift and majority-biased cascades jointly contribute to this generation--retention bottleneck. In biological evolution, selection counteracts stochastic drift by preserving variants with higher fitness. The analogous need in multi-agent reasoning is correctness-sensitive selection pressure: an external LLM judge ranks candidate rationales based on their content rather than answer frequency. This creates a concrete path for a correct but outnumbered candidate to influence the reported answer. The next question is whether an LLM judge can reliably distinguish candidates with correct answers from candidates with incorrect answers, and what factors govern its evaluative fidelity. We therefore map judge reliability across different candidate supply regimes before applying its signal to final answer selection.

{\small\noindent \textbf{a,} Experimental motivation and generation--retention bottleneck. Confounding candidate generation, peer communication, and answer selection obscures the source of multi-agent gains. Inset: Paired accuracy and token cost across four medical reasoning protocols (\emph{n} = 7,350 cases per protocol). The open committee and anonymous evidence board yield marginal accuracy gains (+1.14 pp and +0.83 pp over majority voting) despite using up to 10.73\ensuremath{\times} more tokens. The oracle upper bound reveals a 13.13--14.38 pp generation--retention gap where correct candidate answers are generated initially but not retained in the reported output. \textbf{b,} Conceptual framework of multi-agent information evolution. Candidate generation produces diverse hypotheses; without a reliable quality filter, stochastic variation can give one rationale an early advantage, and repeated peer adoption can reinforce popular errors and crowd out correct minority hypotheses. Memetic drift and majority-biased cascades can therefore jointly create a generation-retention bottleneck. Judge-guided terminal selection provides a content-based signal that can preserve minority correct answers. \textbf{c,} Conceptual decision regime. No selection rule can recover a correct answer when no correct candidate was generated (\emph{p}\textsubscript{gen} = 0, where \emph{p}\textsubscript{gen} is the empirical fraction of correct candidate responses among all valid candidates collected for that question; see Methods, Candidate generation and adaptive candidate banks). When correct answers are present but outnumbered, correctness-sensitive selection has its clearest opportunity to preserve them. At higher availability, majority voting becomes more competitive and the judge's added value declines. Source data are provided with this paper.\par}
\subsection{LLM judge reliability varies with correct-answer availability, task and generator}
The capacity of an LLM judge to provide meaningful selection pressure is not constant, but varies with full-bank correct-answer availability (\emph{p}\textsubscript{gen}) (Prediction 2). To measure this evaluative capacity in isolation from downstream decision rules, we developed an offline ranking benchmark. For each question, repeated generator calls produced a full candidate bank of answers and supporting rationales, defining question-level availability (\emph{p}\textsubscript{gen}) as the empirical fraction of correct candidates (\hyperref[generation]{see Methods, Candidate generation and adaptive candidate banks}). For each sampled pool, a separate ranking-judge call returned an ordering of anonymous candidate rationales. We measured ranking reliability with rank AUC, the probability that a randomly selected correct candidate is ordered above an incorrect candidate from the same pool. A value of 0.5 denotes chance performance (\hyperref[boundary]{see Methods, LLM judge ranking and reliability fitting}).

\begin{figure}[H]
\centering
\phantomsection\label{fig2}
\includegraphics[alt={Fig. 2 | Evaluator ranking reliability varies with correct-answer availability, benchmark and generator setting},width=\linewidth]{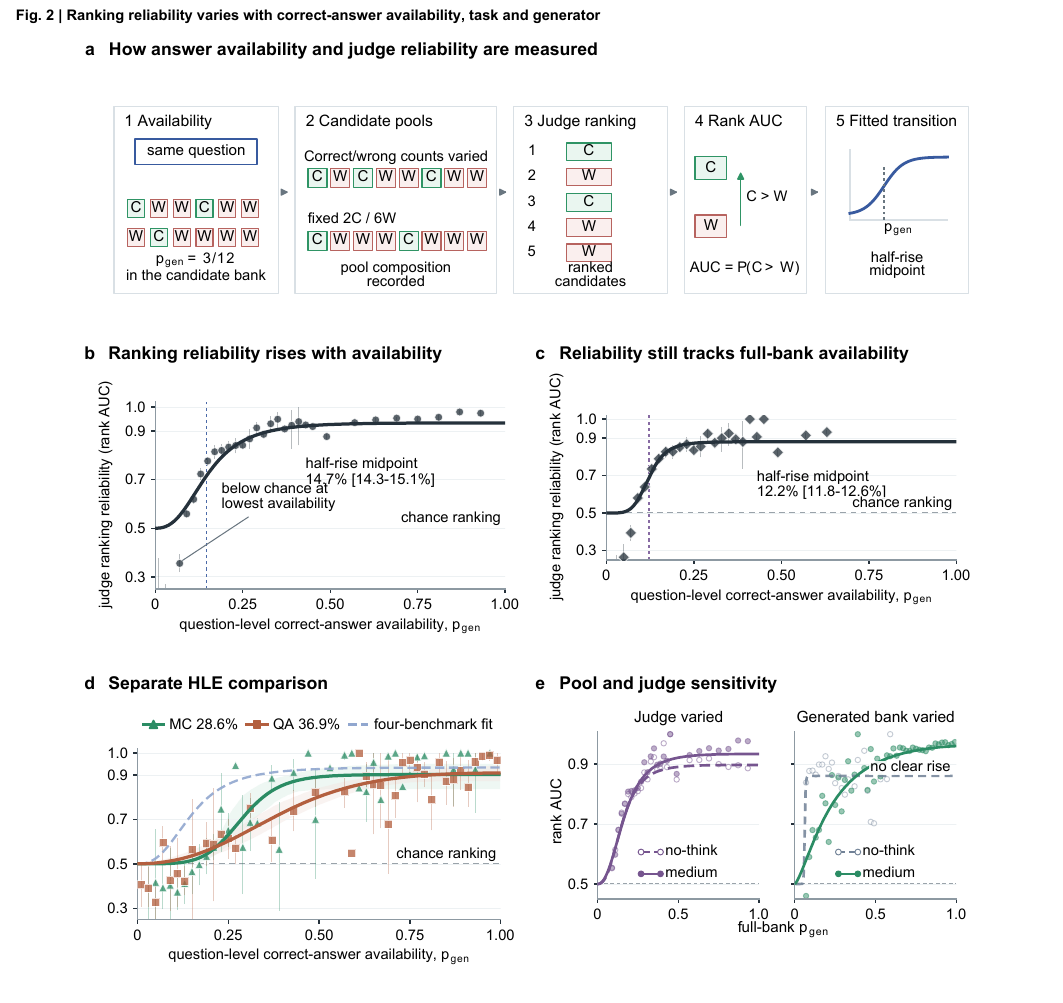}
\caption*{\textbf{Fig. 2 | Evaluator ranking reliability varies with correct-answer availability, benchmark and generator setting}}
\end{figure}

Evaluator ranking reliability rises sharply with full-bank correct-answer availability. In the primary evaluation across four reasoning benchmarks (15,336 questions, 67,163 pools and 823,988 candidate pairs; Extended Data Table 1), ranking fidelity was not an invariant property of the evaluator. In the lowest-availability group, observed rank AUC fell below 0.5 (\hyperref[fig2]{Fig. 2b}), indicating that when candidates with correct answers are sparse, plausible incorrect rationales can dominate the judge's comparisons. As the underlying availability of candidates with correct answers (\emph{p}\textsubscript{gen}) increased, rank AUC rose with it. The fitted curve completed half of its rise above chance at an empirical half-rise midpoint of \emph{p}\textsubscript{gen} = 14.7\% (95\% CI, 14.3--15.1\%; pair-weighted \emph{R}\textsuperscript{2} = 0.900; \hyperref[fig2]{Fig. 2b}). A similar rising relationship appeared when availability was measured from unextended 16-attempt generation quotas (midpoint 13.6\%, 95\% CI, 13.1--14.0\%; \emph{R}\textsuperscript{2} = 0.927; \hyperref[ed_fig2]{Extended Data Fig. 2c}), showing that the relationship was already present in the initial candidate supply.

The association between full-bank availability and judge ranking reliability persists when the displayed correct-to-wrong candidate ratio is held constant. To rule out the visible correct-to-wrong ratio in the prompt as a confounding artifact (i.e., whether the judge merely benefits from seeing more correct options), we evaluated fixed-composition pools containing exactly 2 correct and 6 wrong candidates (2C/6W) across 12,184 questions (146,208 pairs; \hyperref[fig2]{Fig. 2a}). Even when the displayed ratio was held constant, judge reliability continued to track the underlying full-bank availability (\emph{p}\textsubscript{gen}), producing a fitted half-rise midpoint of 12.2\% (95\% CI, 11.8--12.6\%; \emph{R}\textsuperscript{2} = 0.887; \hyperref[fig2]{Fig. 2c}; \hyperref[boundary]{see Methods}). The persistence of this relationship under 2C/6W shows that \emph{p}\textsubscript{gen} is not merely a proxy for the number of correct options displayed to the judge. Under fixed composition, \emph{p}\textsubscript{gen} instead indexes question-level candidate separability: questions that yield few correct traces also tend to produce correct and incorrect rationales that are harder to distinguish. Candidate scarcity, question difficulty and generator-specific candidate content therefore jointly shape the ranking signal.

The fitted availability--reliability relationship shifts toward higher descriptive half-rise midpoints in the HLE evaluations. To test whether the observed reliability boundaries hold under extreme academic reasoning that exceeds standard pretraining knowledge, we extended our evaluation to Humanity's Last Exam (HLE)\cite{ref51}. On HLE multiple-choice questions scored with official keys, the fitted half-rise midpoint shifted upward to 28.6\% (\emph{R}\textsuperscript{2} = 0.810; 475 questions; \hyperref[fig2]{Fig. 2d}). On HLE open-answer questions evaluated via an audited reference-equivalence judge (\hyperref[hle-audit]{see Methods, Separate HLE comparison and open-answer QA audit}), the midpoint shifted further to 36.9\% (\emph{R}\textsuperscript{2} = 0.824; 811 questions; \hyperref[fig2]{Fig. 2d} and \hyperref[ed_fig3]{Extended Data Fig. 3}). These higher descriptive midpoints indicate that the availability associated with reliable ranking shifted upward in the HLE evaluations.

Cross-model evaluations showed that the availability relationship persisted when candidates generated by DeepSeek-V4-Flash (Preview) were ranked by a different judge. When Mimo v2.5 ranked these candidate banks, rank AUC rose strongly with availability under both Mimo no-think (\emph{R}\textsuperscript{2} = 0.922) and medium-thinking (\emph{R}\textsuperscript{2} = 0.915) judge settings (\hyperref[fig2]{Fig. 2e}). No-think means that Mimo's optional internal reasoning was disabled, whereas medium-thinking means that it was enabled at the medium setting (\hyperref[models]{see Methods}).

Changing the candidate generator had a larger effect on the availability--reliability relationship, even with DeepSeek fixed as the judge. Across 12,477 matched benchmark questions, the two Mimo-generated banks yielded identical overall mean rank AUC (0.8493). Despite this identical mean, the availability--reliability patterns differed markedly between the two generator settings. The mean summarizes overall pairwise ranking, whereas \emph{R}\textsuperscript{2} measures how strongly ranking reliability varies with \emph{p}\textsubscript{gen}; the same mean can therefore coexist with different availability-dependent patterns. Mimo no-think candidates showed an almost flat relationship with availability (\emph{R}\textsuperscript{2} = 0.022), whereas Mimo medium-thinking candidates showed a steep sigmoid transition (\emph{R}\textsuperscript{2} = 0.928; half-rise midpoint, 19.0\%; \ensuremath{\Delta}\emph{R}\textsuperscript{2} = 0.905 (95\% CI, 0.865--0.920); \hyperref[fig2]{Fig. 2e} and Extended Data Table 1; \hyperref[matched-generator-bank]{see Methods, Matched generator-bank comparison}). These results show that judge reliability varied across generator, task and candidate regimes and should therefore be evaluated in the intended setting. Together, these ranking benchmarks show that an LLM judge provides an informative correctness signal (rank AUC > 0.5) across moderate and high candidate availability, confirming its technical viability as an external quality filter. However, an above-chance ranking metric (AUC) does not automatically translate into improved system-level accuracy: an evaluator might rank a correct candidate higher than a wrong one without placing it at the very top of the pool or overcoming an overwhelming incorrect majority. We therefore transition from evaluative signal benchmarking to terminal selection, investigating how answer-selection rules can operationalize this judge signal to rescue outnumbered correct candidates within fixed candidate pools.

{\small\noindent \textbf{a,} Evaluation protocol for isolating selection pressure. A generator produces a candidate bank defining question-level availability (\emph{p}\textsubscript{gen}). An LLM judge ranks anonymous rationales without seeing answers or correctness labels. Rank AUC measures the probability of ranking a correct candidate above a wrong one (0.5 = chance). The fitted half-rise midpoint (m) summarizes the availability level at which the fitted curve completes half of its rise above chance. \textbf{b,} Sigmoidal reliability transition. In the primary DeepSeek\ensuremath{\rightarrow}DeepSeek setting across four benchmarks (15,336 questions), observed rank AUC falls below chance in the lowest-availability group and then rises with \emph{p}\textsubscript{gen}; the fitted half-rise midpoint is m = 14.7\% (R\textsuperscript{2} = 0.900). \textbf{c,} Fixed-composition control (2C/6W). Holding the displayed pool at 2 correct and 6 wrong candidates across 12,184 questions preserves the sigmoidal rise (m = 12.2\%, R\textsuperscript{2} = 0.887), showing that \emph{p}\textsubscript{gen} captures question-level candidate separability beyond displayed pool composition. \textbf{d,} HLE comparisons shift the fitted half-rise. On Humanity's Last Exam (HLE), the fitted half-rise midpoints shift to 28.6\% (multiple choice) and 36.9\% (open QA), showing that the amount of correct-answer availability associated with a strong comparative signal differs across benchmark and labelling settings. \textbf{e,} Generator-setting contrast. Across matched questions, Mimo no-think and medium-thinking candidate banks have identical mean rank AUC (0.8493), but their availability--reliability relationships differ: no-think is almost flat (R\textsuperscript{2} = 0.022), whereas medium-thinking shows a sharp sigmoidal transition (R\textsuperscript{2} = 0.928). Source data are provided with this paper.\par}
\subsection{Combining answer frequency and judge rankings recovers outnumbered correct candidates}
\begin{figure}[H]
\centering
\phantomsection\label{fig3}
\includegraphics[alt={Fig. 3 | Terminal selection rules operationalize evaluator selection pressure to recover outnumbered correct answers from frozen candidate pools},width=\linewidth]{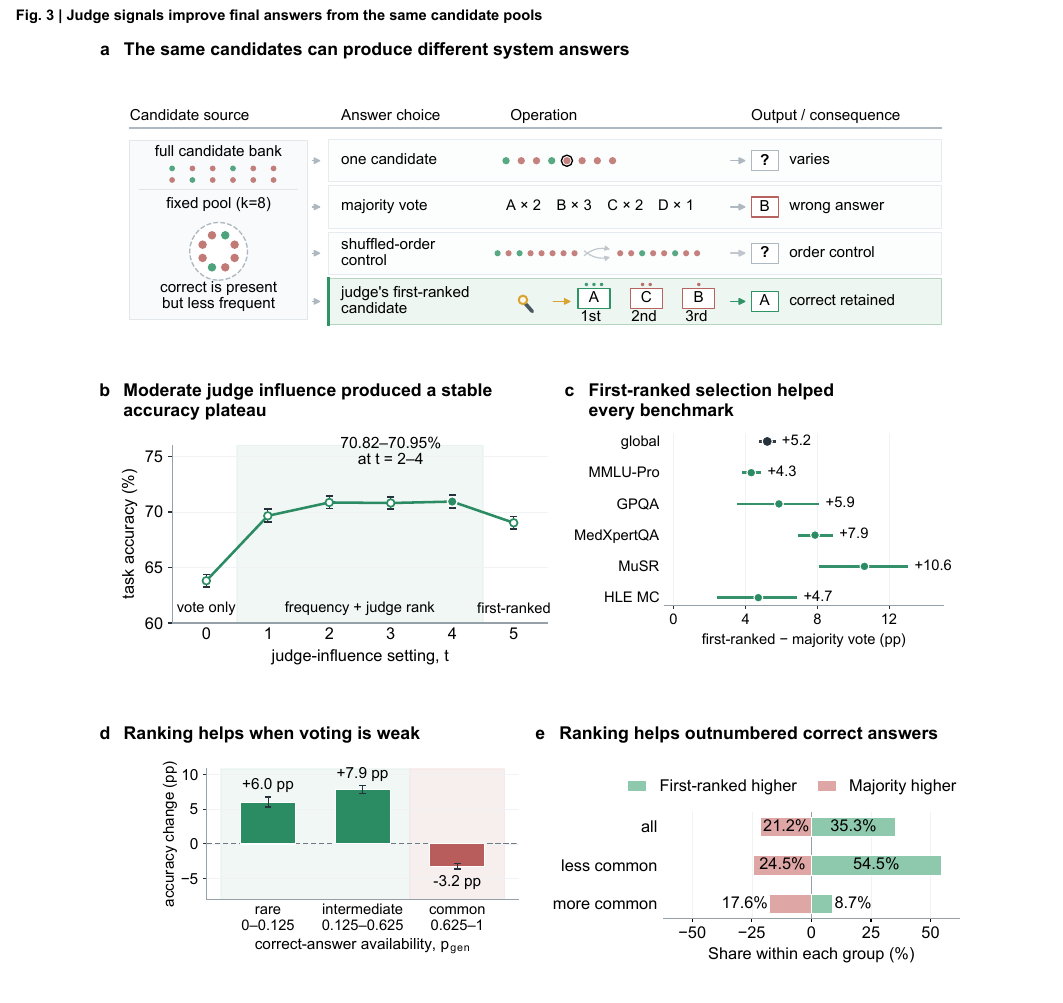}
\caption*{\textbf{Fig. 3 | Terminal selection rules operationalize evaluator selection pressure to recover outnumbered correct answers from frozen candidate pools}}
\end{figure}

Terminal selection rules that combine consensus frequency with judge ranking signals reliably rescue outnumbered correct answers, significantly outperforming pure majority voting on identical candidate pools (Prediction 3). To evaluate how selection pressure alters final answer reporting without the confounding effects of dynamic communication, we conducted offline replays across 81,390 frozen candidate pools (\emph{k} = 8) drawn from 16,278 questions across five benchmarks (\hyperref[fig3]{Fig. 3a} and \hyperref[ed_fig4]{Extended Data Fig. 4a}).

Terminal selection rules that combine consensus frequency with rank-based selection pressure achieve a 70.82--70.95\% accuracy plateau, rescuing outnumbered correct answers without generating new candidate traces in this offline replay. Across 81,390 stored pools of size \emph{k} = 8, majority voting (\emph{t} = 0) achieved 63.82\% accuracy, whereas unguided single-candidate sampling and shuffled-rank controls scored 42.53\% and 42.47\% (\hyperref[fig3]{Fig. 3b} and \hyperref[ed_fig4]{Extended Data Fig. 4b}). Systematically scaling the judge's influence via rank-power weighting (\emph{w}\textsubscript{\emph{i}} = (\emph{k} \ensuremath{-} \emph{i})\textsuperscript{\emph{t}}; \hyperref[replay]{see Methods}) revealed a robust performance plateau at \emph{t} = 2--4 (70.82--70.95\%), outperforming majority voting by 7.00--7.13 percentage points on identical candidate text (\hyperref[fig3]{Fig. 3b}). First-ranked selection (\emph{t} = 5, selecting the top-ranked candidate) achieved 69.04\% (+5.22 pp; 95\% CI, 4.85--5.59; \hyperref[fig3]{Fig. 3b,c}). This \emph{t} = 2--4 plateau reflects a division of labour: answer frequency captures repeated support across independently generated trajectories, while judge ranking can elevate a correct candidate that remains outnumbered. Moderate rank weighting therefore combines the statistical stability of aggregation with a content-sensitive correction for popular errors.

First-ranked selection outperformed majority voting across all five benchmarks, yielding gains of 4.3 percentage points on MMLU-Pro, 5.9 on GPQA, 7.9 on MedXpertQA, 4.7 on HLE MC and 10.6 on MuSR (\hyperref[fig3]{Fig. 3c} and \hyperref[ed_fig4]{Extended Data Fig. 4c}). This performance advantage was largely shaped by correct-answer availability (\emph{p}\textsubscript{gen}). Judge selection improved accuracy by 6.02 percentage points over majority voting when correct candidates were rare (0 < \emph{p}\textsubscript{gen} < 0.125; 95\% CI, 5.31--6.79; \emph{n} = 2,407) and by 7.87 percentage points at intermediate availability (0.125 \ensuremath{\leq} \emph{p}\textsubscript{gen} \ensuremath{\leq} 0.625; 95\% CI, 7.30--8.42; \emph{n} = 10,269; \hyperref[fig3]{Fig. 3d}). By contrast, when correct candidates dominated the candidate bank (0.625 < \emph{p}\textsubscript{gen} < 1), judge selection reduced accuracy by 3.22 percentage points because voting frequency was already highly reliable (\hyperref[fig3]{Fig. 3d}). This dynamic substantially shaped question-level outcomes: in questions where correct candidates were present but less common, first-ranked selection outperformed majority voting in 54.5\% of questions and underperformed it in 24.5\%. Conversely, in majority-correct questions, majority voting was superior in 17.6\% of questions, compared with 8.7\% for first-ranked selection (\hyperref[fig3]{Fig. 3e}).

Judge ranking depends on preserving the correspondence between rationales and their selected conclusions. To determine how candidate information contributes to judge ranking, we performed controlled information perturbations across 713 matched pools where correct answers were outnumbered. Here, \emph{p}\textsubscript{pool} denotes the realised fraction of correct candidates in a sampled pool, unlike \emph{p}\textsubscript{gen}, which measures correct-answer availability in the full candidate bank. The selected pools had 0 < \emph{p}\textsubscript{pool} < 0.5 (\hyperref[prospective]{see Methods, Targeted reranking and information perturbation}; \hyperref[ed_fig5]{Extended Data Fig. 5a}). Presenting unperturbed candidates (rationale plus answer) achieved 0.684 rank AUC. Stripping candidates to answers alone or rationales alone reduced rank AUC to 0.654 and 0.655. Crucially, preserving the original rationale--answer association rather than swapping rationales increased rank AUC by 0.173 (95\% CI, 0.139--0.208; \hyperref[ed_fig5]{Extended Data Fig. 5b,c}); the swapped-rationale condition yielded 0.482 rank AUC. These results show that judge ranking relies on the correspondence between a rationale and its selected conclusion.

Together, these experiments show that terminal selection rules can use judge rankings to rescue minority correct candidates within fixed candidate pools (\emph{k} = 8), substantially outperforming majority voting. However, in practical system design, practitioners face a fundamental trade-off: whether to deploy an external judge on small candidate sets or simply generate more samples to empower majority voting. We evaluate this scaling and cost-efficiency trade-off in the next section.

{\small\noindent \textbf{a,} Mechanisms of terminal answer selection. Combining answer frequency with judge ranking weights (t = 1--4) or using first-ranked selection (t = 5) can select an outnumbered correct candidate (A) over the more frequent incorrect answer (B) in the illustrated fixed pool. \textbf{b,} The hybrid selection plateau. Accuracy across judge influence strength (t = 0--5) over 81,390 stored pools (\emph{n} = 16,278 questions). Majority voting (t = 0) achieves 63.82\%; hybrid rules yield a stable plateau at t = 2--4 (70.82--70.95\%, +7.13 pp gain); first-ranked selection alone (t = 5) achieves 69.04\%. \textbf{c,} Benchmark consistency. First-ranked selection outperforms majority voting across all five benchmarks (+4.33 to +10.62 pp). \textbf{d,} Performance gains by candidate availability. First-ranked selection gains +6.02 to +7.87 pp in rare and intermediate full-bank availability regimes (\emph{p}\textsubscript{gen} \ensuremath{\leq} 0.625), with a -3.22 pp difference at high full-bank availability (\emph{p}\textsubscript{gen} > 0.625). \textbf{e,} Question-level outcome distribution. First-ranked selection is higher in 54.5\% and lower in 24.5\% of questions where correct candidates are present but less common in the full candidate bank; ties are omitted. Source data are provided with this paper.\par}
\subsection{Candidate pool size changes the relative value of voting and judging}
\begin{figure}[H]
\centering
\phantomsection\label{fig4}
\includegraphics[alt={Fig. 4 | Test-time compute scaling shifts the cost--accuracy trade-off between candidate generation volume and evaluator selection pressure},width=\linewidth]{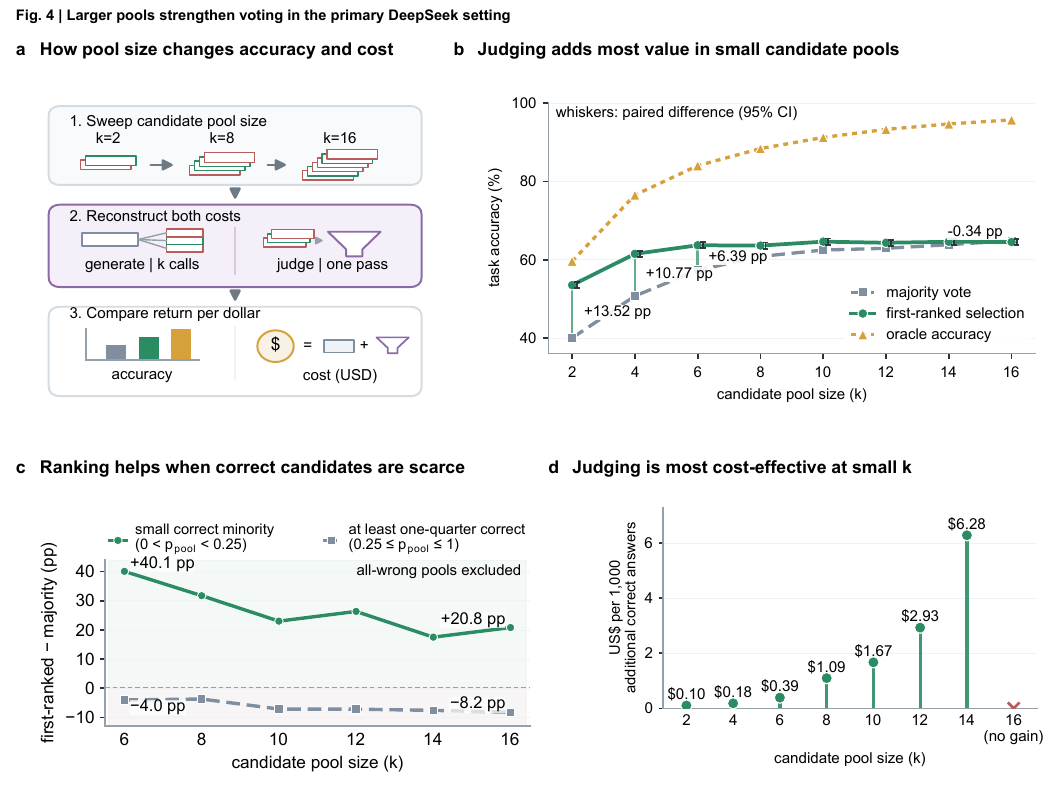}
\caption*{\textbf{Fig. 4 | Test-time compute scaling shifts the cost--accuracy trade-off between candidate generation volume and evaluator selection pressure}}
\end{figure}

\begin{figure}[H]
\centering
\phantomsection\label{ed_fig1}
\includegraphics[alt={Extended Data Fig. 1 | Initial multi-agent experiments show costly communication and loss of available correct answers},width=\linewidth]{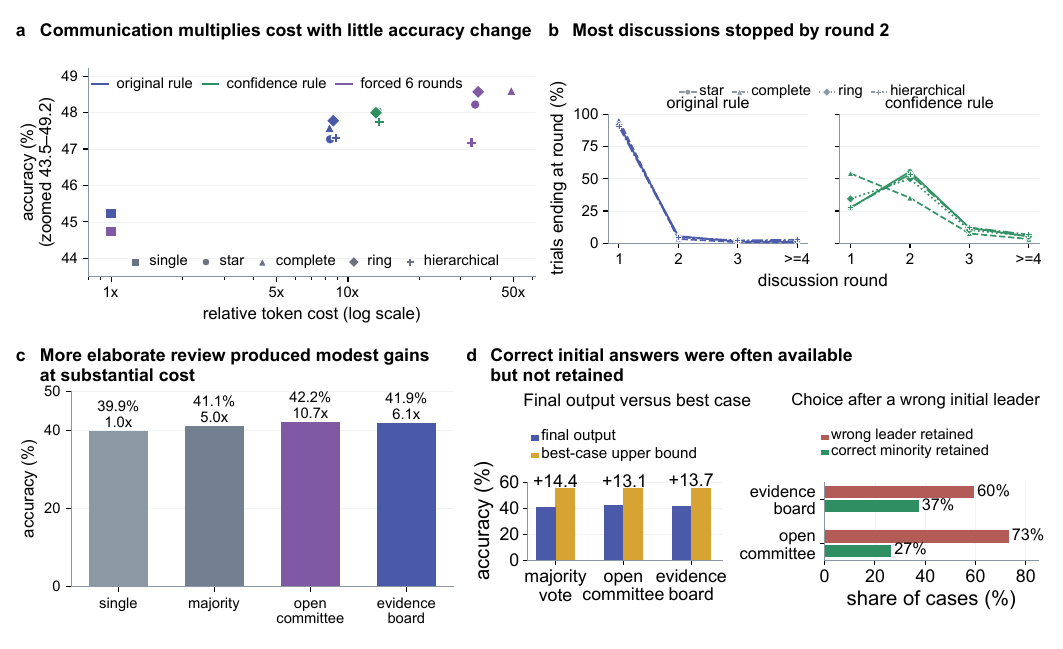}
\caption*{\textbf{Extended Data Fig. 1 | Initial multi-agent experiments show costly communication and loss of available correct answers}}
\end{figure}

\begin{figure}[H]
\centering
\phantomsection\label{ed_fig2}
\includegraphics[alt={Extended Data Fig. 2 | Ranking reliability varies across benchmarks, generators and ranking judges},width=\linewidth]{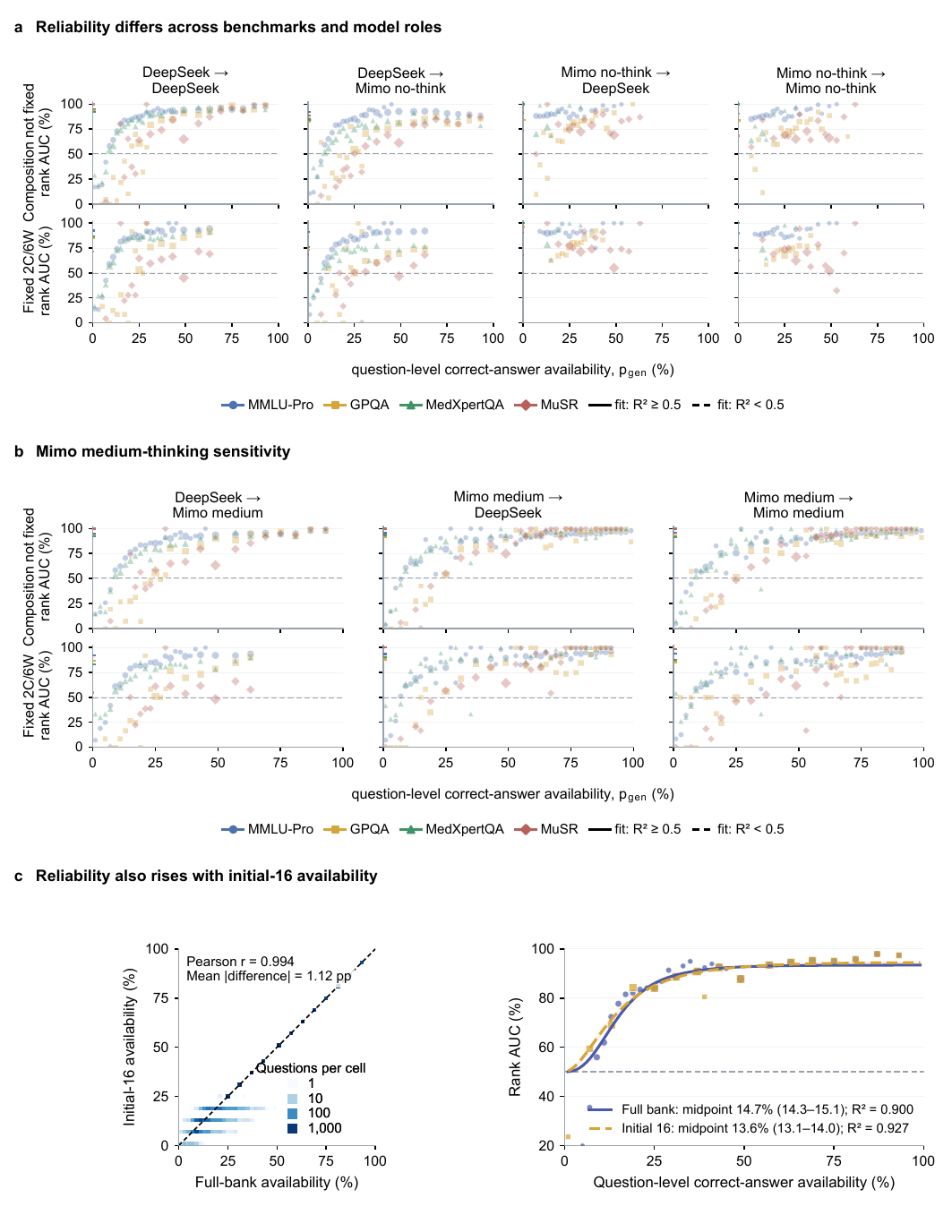}
\caption*{\textbf{Extended Data Fig. 2 | Ranking reliability varies across benchmarks, generators and ranking judges}}
\end{figure}

\begin{figure}[H]
\centering
\phantomsection\label{ed_fig3}
\includegraphics[alt={Extended Data Fig. 3 | Ranking reliability varies across Humanity's Last Exam settings},width=\linewidth]{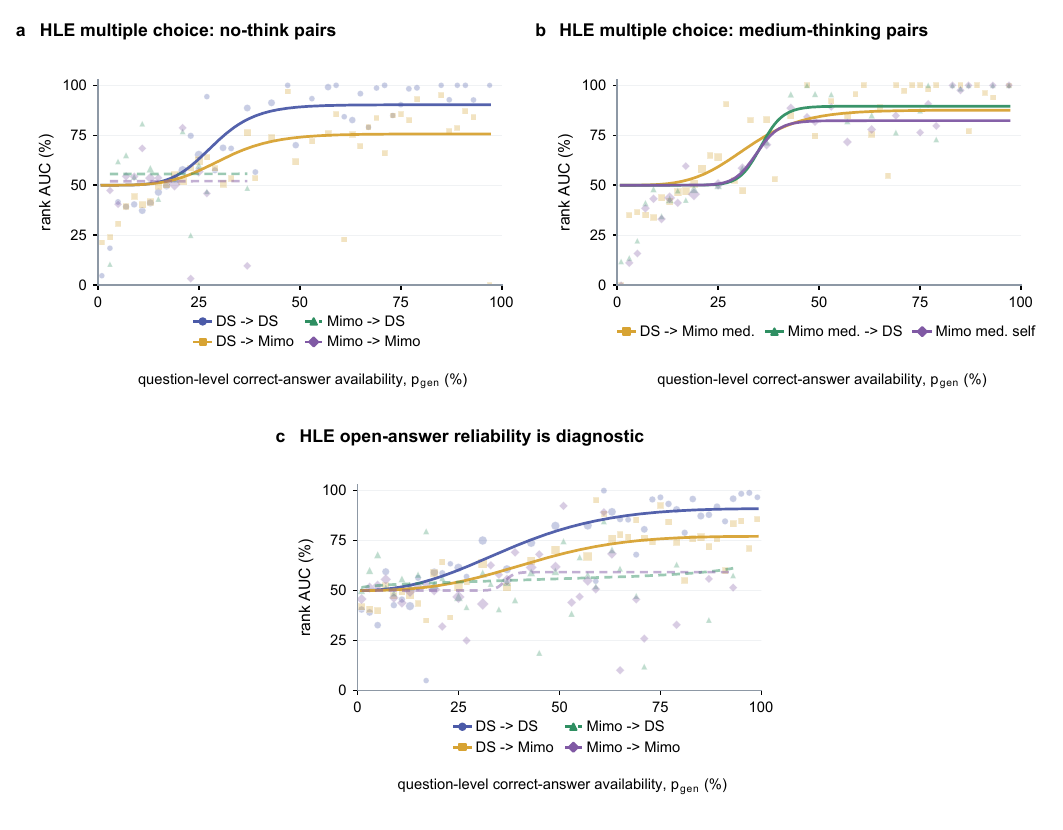}
\caption*{\textbf{Extended Data Fig. 3 | Ranking reliability varies across Humanity's Last Exam settings}}
\end{figure}

\begin{figure}[H]
\centering
\phantomsection\label{ed_fig4}
\includegraphics[alt={Extended Data Fig. 4 | Fixed-pool gains vary by benchmark and correct-answer availability},width=\linewidth]{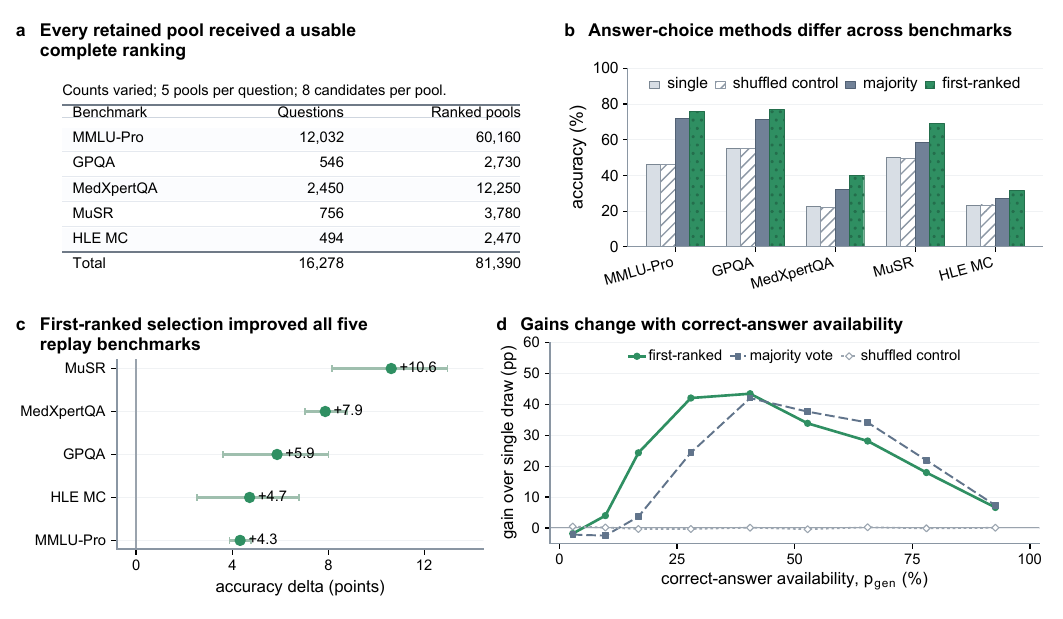}
\caption*{\textbf{Extended Data Fig. 4 | Fixed-pool gains vary by benchmark and correct-answer availability}}
\end{figure}

\begin{figure}[H]
\centering
\phantomsection\label{ed_fig5}
\includegraphics[alt={Extended Data Fig. 5 | Answers and supporting rationales both inform the ranking judge},width=\linewidth]{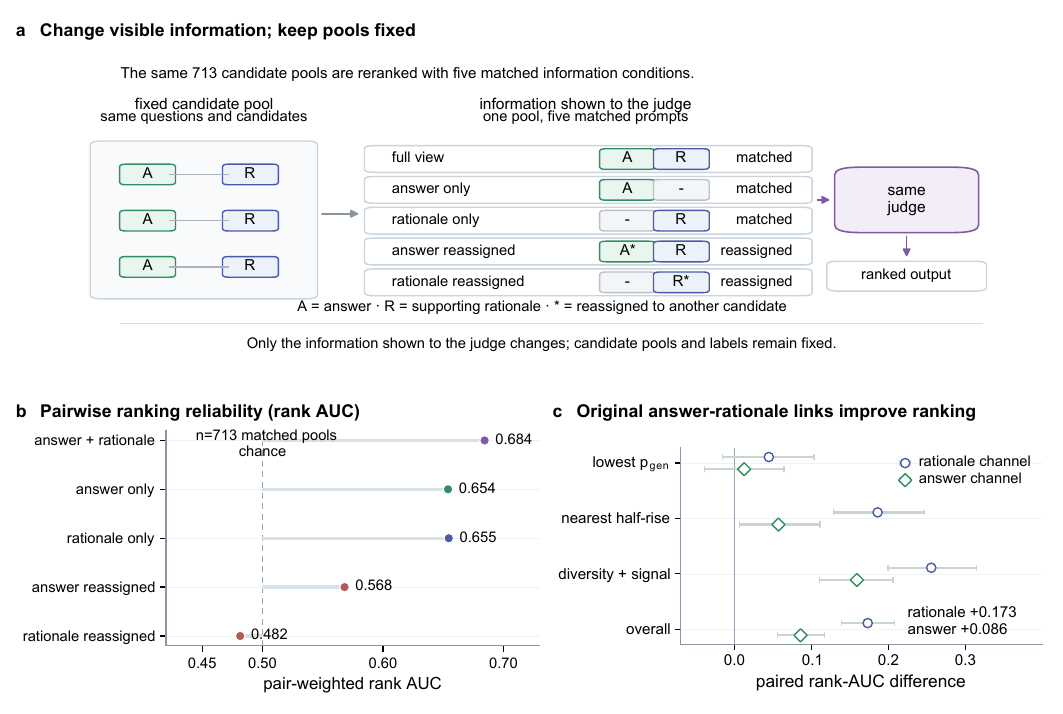}
\caption*{\textbf{Extended Data Fig. 5 | Answers and supporting rationales both inform the ranking judge}}
\end{figure}

Test-time compute scaling alters the trade-off between expanding candidate sample volume and deploying external selection pressure. In practical systems, practitioners must decide whether to allocate test-time compute toward sampling additional candidate trajectories or invoking an evaluator. To map this cost--accuracy trade-off (Prediction 3), we swept candidate pool size (\emph{k}) from 2 to 16 across matched cohorts of up to 11,241 questions, reconstructing token-based generation and judging expenditures under the stated cache assumptions (\hyperref[fig4]{Fig. 4a}; \hyperref[cost]{see Methods, Pool size and reconstructed token-based cost}).

Judge selection delivered its largest accuracy advantage in small candidate pools. At \emph{k} = 2, first-ranked selection outperformed majority voting by 13.52 percentage points (95\% CI, 12.75--14.27). This advantage decreased as pool size increased, although its paired 95\% confidence interval remained above zero through \emph{k} = 12. At \emph{k} = 14 (+0.75 percentage points; 95\% CI, \ensuremath{-}0.05--1.55) and \emph{k} = 16 (\ensuremath{-}0.34 percentage points; 95\% CI, \ensuremath{-}1.15--0.44), the intervals included zero (\hyperref[fig4]{Fig. 4b}).

\begin{figure}[H]
\centering
\phantomsection\label{ed_fig6}
\includegraphics[alt={Extended Data Fig. 6 | Judge-guided gains vary with generation setting and pool size},width=\linewidth]{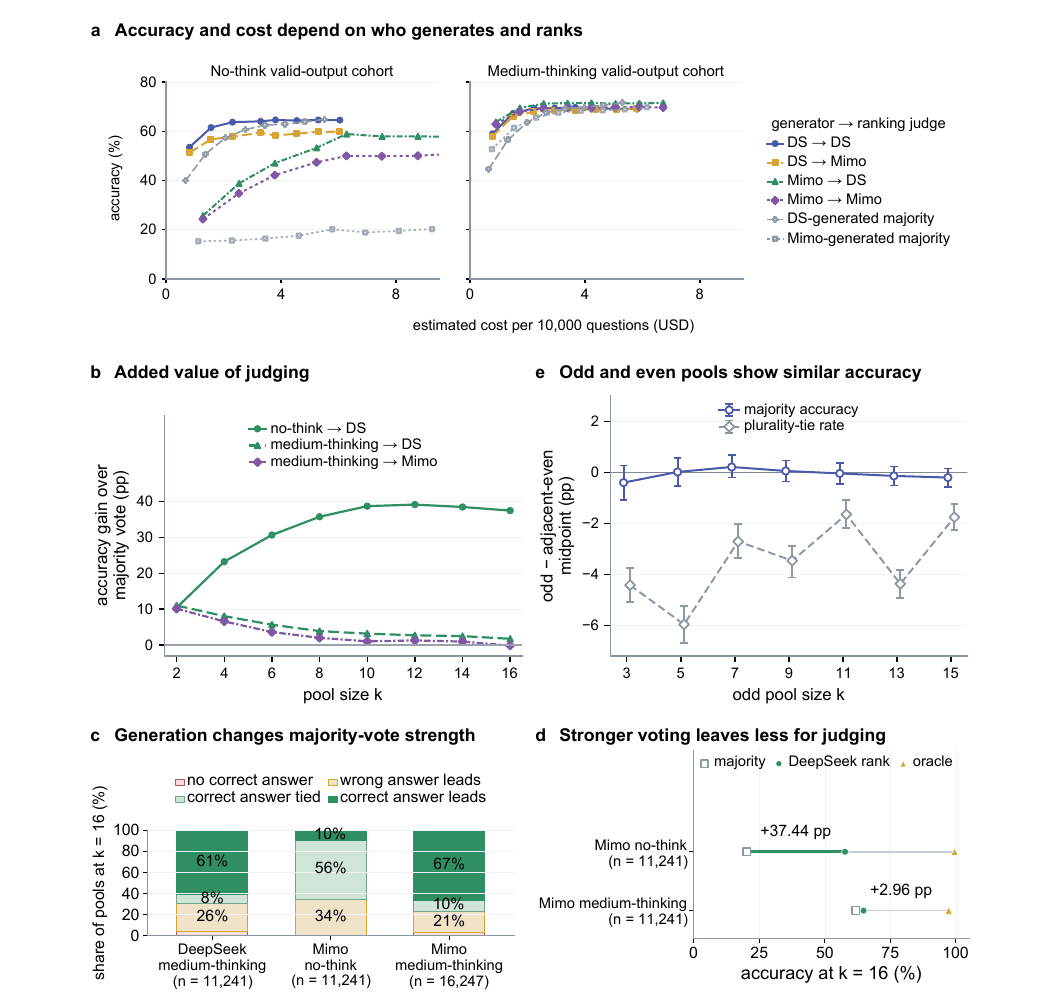}
\caption*{\textbf{Extended Data Fig. 6 | Judge-guided gains vary with generation setting and pool size}}
\end{figure}

The small-pool voting deficit was not explained by plurality ties or the tie-breaking rule. To verify whether even pool sizes unfairly penalise majority voting due to frequent plurality ties (e.g., 1-1 ties at \emph{k} = 2), we evaluated odd pool sizes against adjacent-even midpoints across 11,241 questions. While odd pools reduced plurality tie rates by 1.64--5.96 percentage points, odd-pool voting accuracy differed from adjacent-even midpoints by at most 0.39 pp, with all seven paired 95\% CIs including zero (\hyperref[ed_fig6]{Extended Data Fig. 6e}). Tie-neutral scoring altered accuracy by at most 0.36 pp. Together, these controls rule out tie frequency and tie resolution as the explanation for the small-pool voting deficit.

Stratifying performance by realised pool composition (\emph{p}\textsubscript{pool}) confirmed that judge selection helped most when correct candidates formed a small minority (0 < \emph{p}\textsubscript{pool} < 0.25). All-wrong pools were excluded (\hyperref[fig4]{Fig. 4c}). In these small correct minority pools, judge selection outperformed majority voting by 40.1 percentage points at \emph{k} = 6 and 20.8 percentage points at \emph{k} = 16 (\hyperref[fig4]{Fig. 4c}). By contrast, when correct candidates occupied at least 25\% of the pool (\emph{p}\textsubscript{pool} \ensuremath{\geq} 0.25), majority voting outperformed judge selection by 4.0 percentage points at \emph{k} = 6 and 8.2 percentage points at \emph{k} = 16 (\hyperref[fig4]{Fig. 4c}).

Consequently, judge-guided selection became less cost-effective as candidate volume increased. The pattern reflects a changing source of accuracy: in small pools, individual correct candidates are often outvoted, whereas larger pools allow independently generated correct answers to accumulate enough support for majority voting. In small candidate pools (\emph{k} \ensuremath{\leq} 6), selection pressure was therefore cost-effective, costing US\$0.10 to US\$0.39 per 1,000 additional correct answers over majority voting (\hyperref[fig4]{Fig. 4d}). At larger pool sizes, majority voting became more competitive as correct answers became frequent enough to influence the vote. The incremental cost of evaluator judging escalated to US\$6.28 at \emph{k} = 14 and became economically undefined at \emph{k} = 16 (\hyperref[fig4]{Fig. 4d}), where first-ranked selection no longer improved accuracy over majority voting.

Pool size alone did not determine when majority voting approached judge selection. The generator's reasoning setting also affected how additional samples changed the candidate distribution. On the same 11,241 questions at \emph{k} = 16, the two settings differed sharply. On Mimo no-think pools, majority voting achieved only 20.28\%, but DeepSeek selection boosted it to 57.73\% (+37.44 percentage points; 95\% CI, 36.37--38.49). On Mimo medium-thinking pools, where voting was already strong (61.83\%), selection provided a modest bump to 64.79\% (+2.96 percentage points; 95\% CI, 2.22--3.73; \hyperref[ed_fig6]{Extended Data Fig. 6b,d}). The questions were matched, but each setting generated its own candidate answers and rationales.

These findings support Prediction 3: judge-guided selection yielded its largest gains when correct candidates were present but outnumbered, and its added value shrank when majority voting was already strong. Thus, the benefit of generating more candidates depends on whether they increase the supply of correct candidates, make those candidates frequent enough for voting or leave a minority that an informative judge can recognise.

{\small\noindent \textbf{a,} Token-based cost reconstruction. Schematic illustrating cost allocation: majority voting scales linearly with generation calls (\emph{k}), while first-ranked selection adds one external ranking judge call over the pool. \textbf{b,} Sample scaling frontier. Across matched cohorts of up to 11,241 questions, the first-ranked-minus-majority difference narrows from +13.52 pp at \emph{k} = 2 to +0.75 pp at \emph{k} = 14 and \ensuremath{-}0.34 pp at \emph{k} = 16; capped bars indicate paired 95\% CIs. \textbf{c,} Accuracy differential by realised pool composition (\emph{p}\textsubscript{pool}). In small minority pools (0 < \emph{p}\textsubscript{pool} < 0.25), first-ranked selection gains +40.1 pp at \emph{k} = 6 and +20.8 pp at \emph{k} = 16; in higher-correct-fraction pools (\emph{p}\textsubscript{pool} \ensuremath{\geq} 0.25), majority voting is higher by 4.0 pp at \emph{k} = 6 and 8.2 pp at \emph{k} = 16. \textbf{d,} Marginal cost escalation. The incremental judging cost per 1,000 additional correct answers escalates from US\$0.10 at \emph{k} = 2 to US\$6.28 at \emph{k} = 14, becoming economically undefined at \emph{k} = 16 because first-ranked selection no longer improves accuracy over majority voting. Source data are provided with this paper.\par}
\section{Discussion}
\label{discussion-heading}
This study presents a unified evolutionary framework for explaining when multi-agent language model systems preserve generated correct answers and when they lose them downstream. Rather than treating multi-agent consensus as a single mechanism, our findings show that accuracy in the systems studied here can reflect three separable bottlenecks: whether candidate generation produces a correct answer, whether an LLM judge can recognise that answer among plausible errors (providing selection pressure), and whether the answer-selection rule acts on that signal (terminal selection). This separation accounts for a recurring failure in multi-agent systems: without selection pressure, unfiltered peer interaction can contribute to memetic drift, where groups of agents can generate a correct answer but still fail to report it. In judge-guided systems, final accuracy therefore reflects the combined effects of correct-answer availability, judge reliability and answer selection.

These results also help clarify the roles of test-time computation and inter-agent communication in scaled systems. Recent efforts to scale inference compute often rely on generating more candidate paths or prolonging multi-agent debate\cite{ref14,ref7,ref1,ref52,ref53,ref54}. Across the initial multi-agent experiments, discussions converged rapidly and the tested protocols more often preserved a popular error than an outnumbered correct answer, revealing a generation--retention bottleneck formed jointly by memetic drift and majority-biased cascades\cite{ref28}. Additional discussion rounds substantially increased token use without consistent accuracy gains. Sampling more candidates expands the candidate pool, but extra generation improves the reported answer only if the answer-selection rule can identify the correct candidate or if generation makes that candidate the overwhelming majority. In a separate fixed-pool analysis, judge-guided selection improved accuracy in small candidate pools when correct answers were present but outnumbered.

A central implication of our empirical analysis is that judge reliability varies with the candidates being evaluated rather than being a static model trait. Evaluation performance varies with the candidate generator, task and empirical correct fraction in the relevant candidate bank (\emph{p}\textsubscript{gen}). A judge may rank candidate rationales accurately under one generator while showing a different availability--reliability relationship under another model or reasoning configuration. This difference can occur even on matched questions because each generator setting produces its own candidate traces. This context-dependence cautions against treating static benchmark scores as universal guarantees of evaluator quality. Evaluators may guide agent systems in code synthesis, mathematical verification and complex reasoning\cite{ref35,ref36,ref33,ref34}. Their ability to distinguish correct from wrong candidates should be measured within the generator regime, task context and availability range in which they will operate.

These findings may also apply to compound AI systems that use external verifiers, human feedback or model-based reward signals\cite{ref35,ref36,ref55,ref56,ref57,ref58}. In practice, system designers should adopt a clear diagnostic progression. First, measure whether candidate generation produces a correct option at all. If supply is zero, no downstream selection or communication rule can recover the answer. Second, when correct candidates are available but minority-held, assess evaluator reliability against candidates from that exact generator. Third, compare majority vote, first-ranked judge selection and a small set of pre-specified hybrid rules on held-out data, choosing among them according to accuracy and cost. Hybrid rules can combine complementary evidence. Frequency records repeated support across independently generated candidates, while judging evaluates the structural coherence of individual rationales and their conclusions.

Several open questions define the boundary of these findings and point toward future research. Our offline pool analyses evaluate the final answer-selection step, termed \emph{terminal selection} in the Methods, over fixed candidate traces, isolating it from the dynamic feedback loops that occur during live interaction. In such systems, receiving agents may repeat, ignore, reject or adopt information, changing which candidates remain available. We call this \emph{receiver-mediated local selection}. They may also revise or combine information, which constitutes \emph{transformation} (\hyperref[terminology]{see Methods}). These mid-conversation processes may alter candidate distributions before the system performs terminal selection\cite{ref59,ref60,ref61}. Expanding this framework to track information evolution across dynamic multi-turn graphs, open-ended generation and real-time execution environments represents an important next step. Ultimately, scaling agent systems effectively requires moving beyond simply adding more model instances. By tracking generation, receiver-mediated local selection, transformation and terminal selection, researchers can build transparent, reliable multi-agent architectures that protect correct answers from being lost.

\section{Methods}
\label{methods-heading}
\section{Initial MAS architecture and discussion experiments}
\label{motivation}
To quantify whether multi-agent interaction retains or loses generated correct answers, we evaluated four protocols on the same 2,450 medical reasoning questions from MedXpertQA. We ran each protocol three times, using seed values 42, 43 and 44. This produced 7,350 question evaluations per protocol. Within each run, the selected seed was sent with every model request. The five agents were executed as separate API requests, and each request included a distinct agent identifier in the system prompt. All requests used DeepSeek-V4-Flash (Preview) at temperature 0. Passing a seed and setting temperature to 0 did not guarantee identical outputs because provider execution may remain non-deterministic.

The single-answer protocol generated one independent response per question. The majority-voting protocol sampled five independent responses without peer communication and reported the most frequent answer. The open committee protocol allowed five agents to answer independently in Round 1, then revise their answers in Round 2 after reviewing peer responses, confidence scores and evidence. For the open committee, the final judge received all valid outputs from Rounds 1 and 2 without labels indicating the round. For the anonymous evidence board, a deterministic parser pooled evidence from five independent Round 1 responses, and the final judge received that board without agent identifiers, answer-frequency metadata or explicit links between evidence statements and answer choices.

For the oracle upper bound reported in Fig. 1a, we counted a question as correct whenever at least one initial response contained the correct answer. This quantity measures the maximum answer recoverable from the generated responses rather than the performance of an implementable selection rule.

For the comparisons reported in Fig. 1a, we matched each result from the open committee or evidence board with the majority-voting result for the same question and seed. These matched outcomes formed a paired comparison. To estimate uncertainty in the accuracy difference, we created 10,000 samples from the original set of 2,450 question identifiers. Each sample contained 2,450 identifiers drawn at random with replacement, meaning that the same question could be selected more than once. Whenever a question was selected, we kept all three runs for that question together. We calculated the paired accuracy difference for each sample. After ordering the 10,000 differences from lowest to highest, we used the 2.5th and 97.5th percentiles as the lower and upper limits of the 95\% confidence interval. An interval that included zero was compatible with no accuracy difference between the two protocols. The confidence-interval calculation used fixed random seeds: 46 when comparing the open committee with majority voting and 49 when comparing the evidence board with majority voting. These seeds controlled only the confidence-interval calculation, not the model runs.

To evaluate whether prolonging multi-agent discussion resolves answer loss, we conducted a communication topology sweep across four five-agent network layouts (Extended Data Fig. 1): star, in which peripheral agents communicate through a central agent A0; complete, in which all agents observe all peer messages; ring, in which agents observe only immediate left and right neighbours; and hierarchical, in which leaf agents report through intermediate agents to a root. Extended Data Fig. 1 reanalysed the preserved outputs from this sweep. The original stopping rule ended a run when all agents agreed or when a confidence-weighted leading answer met the configured weight and mean-confidence criteria. Confidence-qualified consensus required the confidence criterion even when all agents agreed, whereas forced-six-round execution disabled early stopping. Each topology-by-stopping-condition cell contained 3,150 trials. Token consumption was recorded across all API requests and normalised relative to the single-agent baseline.

\section{Candidate generation and adaptive candidate banks}
\label{generation}
A candidate trace comprises one generated answer (an option for closed-choice tasks or a free-form response for HLE QA) together with its supporting rationale. Candidate traces were generated independently by prompting endpoint models (DeepSeek-V4-Flash (Preview) and Mimo v2.5) across multiple-choice reasoning benchmarks (MMLU-Pro, GPQA, MedXpertQA, MuSR and HLE MC) and open-answer QA (HLE QA). Inference conditions included "no-think" (disabling optional internal reasoning) and "medium-thinking" (enabling provider internal reasoning at medium effort; Extended Data Table 2). For benchmarks scored with official answer keys, a correctness label indicates whether the selected answer matches the key; it does not establish that the supporting rationale is free of logical errors.

To encourage rationale diversity across plausible alternatives during closed-choice generation, each prompt included a sample index that cycled a rotating focus option through available choices (A, B, C, etc.). The prompt asked the model to consider that choice while remaining free to select a different option if the focus option could not be defended.

All parse-valid traces collected for a question formed its full candidate bank. To ensure candidate banks contained sufficient correct and wrong candidates for ranking analyses across difficult questions, generation followed an adaptive collection quota. Each question was allocated 16 initial attempts. If the bank contained at least 2 correct and 6 wrong traces, collection stopped. If no correct candidate was generated, the budget expanded up to 128 attempts; for all other quota deficiencies (1 correct trace or fewer than 6 wrong traces), the budget expanded up to 64 attempts. Question-level correct-answer availability, \emph{p}\textsubscript{gen}, is the empirical fraction of correct parse-valid traces in the question's full candidate bank under this collection protocol:

\[
p_{\mathrm{gen}}(q) =
\frac{\text{number of parse-valid traces with a correct answer for }q}
{\text{total number of parse-valid traces for }q\text{ in the candidate bank}}
\]
A closed-choice response was parse valid only when the option mapped to a legal choice and a non-empty rationale was present. Generation parse failures were excluded from candidate-bank denominators and pool construction, while ranking parse failures were excluded from parse-valid ranking analyses. Candidate banks that remained quota-deficient were retained for availability summaries but excluded from analyses whose pool construction required the missing candidate class or count. Because questions lacking required traces received additional sampling attempts, \emph{p}\textsubscript{gen} represents the empirical correct fraction collected under this protocol rather than an unconstrained single-call success probability.

For the initial-16 sensitivity analysis, we also calculated the fraction of correct traces among parse-valid traces from each question's first 16 generation attempts. We refer to this measure as initial-16 availability. As in the full-bank calculation, an attempt that did not yield a parse-valid trace was excluded from both the numerator and denominator.

\section{LLM judge ranking and reliability fitting}
\label{boundary}
To measure whether an LLM judge can place candidates with correct answers above candidates with incorrect answers, candidate pools of size \emph{k} = 8 were drawn from each question's candidate bank and presented in a separate call to the ranking judge at temperature 0. The ranking judge received the question identifier, category, question, answer options and eight anonymous rationales in shuffled order, but not the candidates' selected answers, correctness labels or generation metadata. A ranking was usable only when it contained every anonymous candidate identifier exactly once.

Two candidate pool sampling designs were evaluated:

\begin{itemize}
\item \textbf{Pools sampled without fixed composition:} Each eligible question contributed five pools sampled without replacement from its parse-valid candidate bank using global seed 0. Questions with fewer than eight parse-valid traces were excluded from this design.
\item \textbf{Fixed-composition pools (2C/6W control):} Each eligible question contributed one pool containing exactly 2 correct and 6 wrong candidate traces. This held displayed composition constant while characterising the association between full-bank availability (\emph{p}\textsubscript{gen}) and rank AUC.
\end{itemize}
Judge reliability was quantified using rank AUC, the probability that a randomly selected correct candidate is ranked above a wrong candidate within the pool. For each 0.02-wide bin of full-bank availability (\emph{p}\textsubscript{gen}), pair-weighted rank AUC was computed across all valid correct--wrong candidate pairs. The relationship between correct-answer availability and judge reliability was fitted using a three-parameter sigmoid function via nonlinear weighted least squares:

\[
\operatorname{AUC}(p) = 0.5 + A\,\operatorname{sigmoid}
\!\left[\beta\left\{\operatorname{logit}(p)-\operatorname{logit}(m)\right\}\right]
\]
where sigmoid(\emph{x}) = 1/(1 + exp(-\emph{x})), logit(\emph{p}) = log[\emph{p}/(1 - \emph{p})], \emph{A} is the fitted rise above chance (0.5), \emph{\ensuremath{\beta}} controls curve steepness, and \emph{m} is the half-rise midpoint, the availability level (\emph{p}\textsubscript{gen}) at which the fitted curve completes half of its rise above chance. Parameters were constrained to \emph{A} \ensuremath{\in} [0, 0.55], \emph{\ensuremath{\beta}} \ensuremath{\in} [0.01, 20] and \emph{m} \ensuremath{\in} [0.001, 0.999], with at most 20,000 optimizer evaluations. The fitting scale was 1/sqrt(pair count), giving each bin an effective least-squares weight proportional to its correct--wrong pair count. All populated bins entered the fit; the non-negative \emph{A} constraint fixed the lower asymptote at 0.5. We interpreted half-rise midpoints for fits with weighted \emph{R}\textsuperscript{2} \ensuremath{\geq} 0.50. Before fitting any curve, observations with \emph{p}\textsubscript{gen} = 0 or \emph{p}\textsubscript{gen} = 1 were assigned to the first or last 0.02-wide bin, respectively. The fit used the bin midpoints, which lie between 0 and 1 and therefore have defined logit values. Uncertainty was estimated via 1,000 benchmark-stratified question-bootstrap resamples with seed 20260710.

Across MMLU-Pro, GPQA, MedXpertQA and MuSR, 15,784 questions had candidate banks. Of these, 15,336 contributed at least one ranked pool containing both correct and incorrect candidates and were therefore eligible for the primary rank-AUC analysis.

The initial-16 analysis reused the same 15,336 questions, 67,163 candidate pools, 823,988 correct--wrong candidate comparisons and judge rankings as the primary DeepSeek-generated and DeepSeek-ranked analysis. We grouped these observations into 0.02-wide bins using initial-16 availability and repeated the same curve fit.

The primary 2C/6W fixed-composition control included 12,184 questions, with one pool per question (12,184 pools and 146,208 correct--wrong pairs).

\subsection{Separate HLE comparison and open-answer QA audit}
\label{hle-audit}
Humanity's Last Exam (HLE) was analysed separately from the primary four-benchmark fit. HLE multiple-choice used official benchmark keys; 475 questions contributed to the fit and 494 to replay. For HLE QA, a temperature-zero DeepSeek equivalence judge received the question, reference answer and candidate answer and returned correct, wrong or uncertain; wrong and uncertain were both scored as non-correct. The later ranking judge received neither the reference answer nor the equivalence label. Each generator-specific candidate-bank set contained 1,645 questions before pool eligibility was applied; the main HLE QA fit included 811 questions that contributed at least one ranked pool containing both correct and wrong traces.

To assess the equivalence labels, one author manually reviewed 127 generated candidate answers. Three were too ambiguous to label reliably. Among the remaining 124 answers, the model's label agreed with the human assessment in 120 cases (96.8\%), and all 124 outputs followed the required response format. We also tested 200 constructed controls with known labels: 100 equivalent to the reference answer and 100 non-equivalent. The model labelled 195 controls correctly (97.5\%), and 197 outputs followed the required response format.

\subsection{Matched generator-bank comparison}
\label{matched-generator-bank}
To compare Mimo no-think and medium-thinking generation under the same DeepSeek judge, questions were matched across benchmark identifiers (12,477 questions without fixed pool composition; 7,544 questions for 2C/6W fixed composition). Each generator bank retained its own generated candidate traces and full-bank \emph{p}\textsubscript{gen}. Paired curve differences used 1,000 benchmark-stratified question bootstraps (seed 20260710).

\section{Offline replay over fixed candidate pools}
\label{replay}
Global replay evaluated answer-selection rules over 81,390 pre-computed \emph{k} = 8 candidate pools sampled without fixed composition from 16,278 benchmark questions (12,032 MMLU-Pro, 546 GPQA, 2,450 MedXpertQA, 756 MuSR and 494 HLE MC; 5 pools per question). Candidate text, rationales and answer frequencies were fixed. Stored judge orderings were also fixed except in the prespecified shuffled-order control.

For the single-candidate baseline, one candidate trace was sampled independently from the full parse-valid candidate bank for each seed and was not restricted to the corresponding eight-trace pool. The shuffled-order control replaced the stored judge ordering with a seed-specific random permutation of the same eight candidate identifiers, then applied the same rank-based answer-selection procedure. Single-candidate sampling, majority-vote tie resolution and the shuffled-order control were each evaluated with deterministic seeds 0--99. Replay scores for pool-based strategies were averaged over seeds and retained pools for each question before calculating global and benchmark means. Global uncertainty used 1,000 question-level bootstrap resamples (seed 20260628).

For Fig. 3e, we compared first-ranked selection and majority voting at the question level after averaging accuracy across the same five stored pools of eight candidates. Each question was classified as first-ranked higher, majority higher or tied. We calculated the proportion of these outcomes across all questions and separately when correct candidates were present but less common in the full candidate bank (0 < \emph{p}\textsubscript{gen} < 0.5) or more common (0.5 \ensuremath{\leq} \emph{p}\textsubscript{gen} < 1). Questions with \emph{p}\textsubscript{gen} = 0 or 1 were included in the overall comparison; all 363 were ties and were therefore not displayed as a separate group.

\subsection{Ranking influence parameter (\emph{t})}
\label{replay}
We evaluated the complete sequence \emph{t} = 0--5 as a sensitivity analysis of how strongly the judge's ordering influenced answer selection relative to candidate frequency. The \emph{t} = 2--4 plateau reported in the Results was the flat region observed in this sequence. A trace at zero-based rank \emph{i} received weight \emph{w}\textsubscript{\emph{i}} = (\emph{k} - \emph{i})\textsuperscript{\emph{t}}. For example, when \emph{k} = 8 and \emph{t} = 2, ranks 0--7 received weights 64, 49, 36, 25, 16, 9, 4 and 1. We summed the weights of traces supporting the same answer and reported the answer with the highest total. Tied answer scores were resolved separately with deterministic seeds 0--99, and the reported replay score was the average across these 100 results. The setting \emph{t} = 0 reproduces majority voting and \emph{t} = 1--4 combines frequency with rank ordering. At \emph{k} = 8 and \emph{t} = 5, 8\textsuperscript{5} = 32,768 > \ensuremath{\sum}\textsubscript{\emph{j}=1}\textsuperscript{7} \emph{j}\textsuperscript{5} = 7\textsuperscript{5} + 6\textsuperscript{5} + 5\textsuperscript{5} + 4\textsuperscript{5} + 3\textsuperscript{5} + 2\textsuperscript{5} + 1\textsuperscript{5} = 29,008. The top-ranked trace therefore outweighs all seven lower-ranked traces combined, ensuring that the first-ranked candidate determines the answer.

\section{Targeted reranking and information perturbation}
\label{prospective}
Targeted reranking evaluated 713 matched eight-trace pools containing outnumbered correct candidates, where \emph{p}\textsubscript{pool} is the realised fraction of correct traces in a sampled pool and 0 < \emph{p}\textsubscript{pool} < 0.5. The pools came from three generator--judge cells: DeepSeek\ensuremath{\rightarrow}DeepSeek on HLE MC, DeepSeek\ensuremath{\rightarrow}DeepSeek on GPQA and Mimo no-think\ensuremath{\rightarrow}DeepSeek on GPQA.

Within each cell, we formed three non-overlapping operational samples, each capped at 80 questions with at least 16 traces in the full candidate bank: the lowest \emph{p}\textsubscript{gen} values, the questions nearest the fitted half-rise and the highest diversity-plus-signal scores. The diversity-plus-signal score was \emph{p}\textsubscript{gen}(1 - \emph{p}\textsubscript{gen}) max(fitted AUC - 0.5, 0).

Candidate pools were reranked under five visible information conditions: (1) answer plus rationale, (2) answer only, (3) rationale only, (4) answers reassigned across candidate labels, and (5) rationales reassigned across candidate labels. Reassignments used deterministic within-pool permutations, rotating identity permutations by one position to ensure content moved. Pair-weighted rank AUC was computed across matched pools (6,488 correct--wrong comparisons per condition). Paired condition differences used 2,000 pool-level bootstrap resamples (seed 1729).

\section{Pool size and reconstructed token-based cost}
\label{cost}
The primary pool-size analysis evaluated \emph{k} \ensuremath{\in} \{2, 4, 6, 8, 10, 12, 14, 16\}. At each \emph{k}, we retained questions with parse-valid outputs for every strategy compared at that pool size. The primary DeepSeek (medium-thinking) cohorts contained 11,207--11,241 questions (\ensuremath{\geq}99.7\% of the largest cohort). The separate Mimo medium-thinking cohort at \emph{k} = 16 contained 16,247 questions. The direct no-think versus medium-thinking comparison at \emph{k} = 16 used the same 11,241 question identifiers, with Mimo generating the candidates and DeepSeek ranking them.

At each pool size \emph{k}, majority voting and first-ranked selection used exactly the same \emph{k} candidate traces. The ranking judge ordered that \emph{k}-candidate pool, and first-ranked selection returned the answer attached to the first-ranked trace.

For the primary DeepSeek pool-size comparison and each Mimo generation setting in the question-matched \emph{k} = 16 comparison, we used 10,000 benchmark-stratified paired-question bootstrap resamples. Within each benchmark and generation setting, questions were sampled with replacement, while majority-voting and DeepSeek first-ranked-selection outcomes remained paired. We calculated the accuracy difference in each resample and used the 2.5th and 97.5th percentiles as the limits of the 95\% confidence interval (seeds: 20260810 for DeepSeek and 20260710 for Mimo).

\subsection{Odd-versus-even pool-size sensitivity analysis}
\label{odd-pool-sensitivity}
Each preserved \emph{k} = 16 DeepSeek candidate pool sampled without fixed composition was truncated to every pool size from \emph{k} = 2 to 16 across the same 11,241 questions. Tied plurality winners were selected uniformly using deterministic seed 20260701. A tie-neutral sensitivity instead averaged correctness across tied winners; accuracy changed by at most 0.36 percentage points across \emph{k} (0.01 percentage points at \emph{k} = 2). For each odd \emph{k}, the adjacent-even midpoint was the arithmetic mean of the results at \emph{k} - 1 and \emph{k} + 1. Majority-vote accuracy and plurality-tie rate were compared with this midpoint. Two-sided 95\% confidence intervals used 1,000 paired question bootstraps (seed 20260722). This offline analysis made no model calls and did not reuse the inherited \emph{k} = 16 judge ranking at smaller pool sizes.

\subsection{Token cost reconstruction and blended pricing}
\label{cost}
Costs were reconstructed from stored character counts. For each prompt message, estimated tokens equalled ceil(character count/4) + 12; structured completion tokens equalled ceil(serialized JSON character count/4). Generation prompts, generation completions, ranking prompts and ranking completions were reconstructed separately. DeepSeek pricing used US\$0.0028 per 1M cache-hit input tokens, US\$0.14 per 1M cache-miss input tokens and US\$0.28 per 1M output tokens. Mimo rates of CNY 0.02, CNY 1.00 and CNY 2.00 per 1M tokens respectively denoted cache-hit input, cache-miss input and output, and were converted at CNY 7.25 per US dollar. Input cost used a 90\% cache-hit and 10\% cache-miss blend. The incremental judging cost per 1,000 additional correctly answered questions was calculated as:

\[
\begin{aligned}
&\text{incremental judging cost per 1,000 additional correctly answered questions}\\
&\qquad = 1{,}000\times
\frac{C_{\mathrm{rank}}-C_{\mathrm{majority}}}
{N_{\mathrm{rank}}-N_{\mathrm{majority}}}.
\end{aligned}
\]
where \emph{C} and \emph{N} denote total reconstructed USD cost and number of correctly answered questions for the respective strategies across matched cohort prompts. This quantity is defined only when \emph{N}\textsubscript{rank} > \emph{N}\textsubscript{majority}.

\section{Statistical analysis, exclusions and data-quality checks}
\label{statistics}
All confidence intervals were two-sided percentile 95\% bootstrap intervals. Resampling followed the experimental unit: questions for question-level comparisons and matched pools for information-perturbation contrasts. Replicate counts and seeds are reported in the corresponding subsections. Figure 4b reports paired 95\% confidence intervals for first-ranked selection relative to majority voting; Fig. 4c and d report descriptive point estimates.

We excluded generation or ranking outputs that did not parse and candidate banks that could not supply the traces required by the stated pool design. Pools containing only correct or only wrong traces were excluded from rank AUC calculations because they contained no correct--wrong comparisons. The three ambiguous HLE QA audit cases were excluded from the human--model agreement calculation. In the cost-analysis audit across 48 primary model-and-pool-size combinations, 308,456 of 309,612 tasks produced parse-valid outputs (99.63\%).

\section{Software, provenance and reproducibility}
\label{reproducibility}
The analysis code targets Python 3.10 or later (manuscript build verified under Python 3.12 using NumPy, pandas, SciPy, Matplotlib and statsmodels). The released implementation separates candidate-bank construction, judge ranking, curve fitting, fixed-pool replay, candidate-information perturbation and pool-size cost analysis into versioned modules controlled by configuration manifests.

Analysis scripts and reproduction manifests are versioned at GitHub (\url{https://github.com/YSTLab/mas-reasoning}, release commit \texttt{7840cf3d\allowbreak{}bde77e61\allowbreak{}a1b4349a\allowbreak{}d3239669\allowbreak{}a7502431}). The complete access-controlled research archive (containing raw model outputs, candidate banks, ranked pools, and analysis summaries) is deposited in the Hugging Face dataset repository (\url{https://huggingface.co/datasets/Biogod/mas_reasoning}, revision \texttt{60a16ff6\allowbreak{}026f96d6\allowbreak{}b9da7881\allowbreak{}6d60eeeb\allowbreak{}14d43fdf}). No human participant data were collected.

\section{Definitions and notation}
\label{terminology}
During live multi-agent communication, agents generate, observe, and aggregate responses. A receiving agent may repeat, ignore, reject or adopt peer information (\textbf{receiver-mediated local selection}) or revise and combine statements into a new rationale (\textbf{transformation}). Across rounds, stochastic variation in which statements persist constitutes \textbf{memetic drift}, whereas repeated adoption of an already popular statement can produce a \textbf{majority-biased cascade}. These processes can coexist and jointly create a generation--retention bottleneck. Together, they form part of the \textbf{information evolution} framework discussed in the main text. The fixed-pool replay experiments evaluate judge ranking and \textbf{terminal selection}, the final answer-selection step, over stored candidate traces. By holding those traces and their judge rankings fixed, the replay isolates terminal selection from the live communication processes that repeat, reject, adopt, revise or combine information.

\section{Models and inference conditions}
\label{models}
We tested seven generator--judge combinations. DeepSeek-V4-Flash (Preview), accessed through the API in June and July 2026, used medium-thinking for the reported generation and ranking conditions. Mimo v2.5 was evaluated with optional internal reasoning either disabled (no-think) or enabled at the medium setting (medium-thinking). Across the candidate-bank and ranking experiments, generation used temperature 1.0 and ranking used temperature 0.0. Extended Data Table 2 lists all seven generator--judge combinations and the corresponding reasoning controls.

\begin{landscape}
\begingroup
\scriptsize
\setlength{\tabcolsep}{1.4pt}
\renewcommand{\arraystretch}{1.15}
\phantomsection\label{ed_table_1}
\noindent \textbf{Extended Data Table 2 | Tested generator and ranking-judge combinations} Generator models, ranking-judge models, sampling temperatures and reasoning controls across all seven tested combinations.\par\medskip
\begin{longtable}{@{}>{\raggedright\arraybackslash}p{0.1068\linewidth}>{\raggedright\arraybackslash}p{0.1602\linewidth}>{\raggedright\arraybackslash}p{0.1157\linewidth}>{\raggedright\arraybackslash}p{0.1335\linewidth}>{\raggedright\arraybackslash}p{0.0890\linewidth}>{\raggedright\arraybackslash}p{0.2848\linewidth}@{}}
\toprule
Combination label & Generator model & Ranking-judge model & Generation temp. & Judge temp. & Reasoning control \\
\midrule
\endfirsthead
\toprule
Combination label & Generator model & Ranking-judge model & Generation temp. & Judge temp. & Reasoning control \\
\midrule
\endhead
DeepSeek (medium-thinking) \ensuremath{\rightarrow} DeepSeek (medium-thinking) & DeepSeek-V4-Flash (Preview) & DeepSeek-V4-Flash (Preview) & 1.0 & 0.0 & Medium-thinking enabled for generation and ranking. \\
DeepSeek (medium-thinking) \ensuremath{\rightarrow} Mimo (no-think) & DeepSeek-V4-Flash (Preview) & Mimo v2.5 & 1.0 & 0.0 & Medium-thinking generator; no-think judge. \\
DeepSeek (medium-thinking) \ensuremath{\rightarrow} Mimo (medium-thinking) & DeepSeek-V4-Flash (Preview) & Mimo v2.5 & 1.0 & 0.0 & Medium-thinking enabled for generation and ranking. \\
Mimo (no-think) \ensuremath{\rightarrow} DeepSeek (medium-thinking) & Mimo v2.5 & DeepSeek-V4-Flash (Preview) & 1.0 & 0.0 & No-think generator; medium-thinking judge. \\
Mimo (no-think) \ensuremath{\rightarrow} Mimo (no-think) & Mimo v2.5 & Mimo v2.5 & 1.0 & 0.0 & Optional internal reasoning disabled throughout. \\
Mimo (medium-thinking) \ensuremath{\rightarrow} DeepSeek (medium-thinking) & Mimo v2.5 & DeepSeek-V4-Flash (Preview) & 1.0 & 0.0 & Medium-thinking enabled for generation and ranking. \\
Mimo (medium-thinking) \ensuremath{\rightarrow} Mimo (medium-thinking) & Mimo v2.5 & Mimo v2.5 & 1.0 & 0.0 & Medium-thinking enabled for generation and ranking. \\
\bottomrule
\end{longtable}
\endgroup
\end{landscape}
\section{Benchmarks and correctness labels}
\label{benchmarks}
The primary modelling set comprised MMLU-Pro, GPQA, MedXpertQA and MuSR, using the stored benchmark splits and official answer keys. HLE multiple-choice questions used official keys and were analysed as a separate external comparison. HLE open-answer questions used labels from the separate reference-answer equivalence judge described above and therefore remained diagnostic evidence. Extended Data Table 3 gives the stored source, split, correctness source and study role for each benchmark.

\begingroup
\scriptsize
\setlength{\tabcolsep}{1.4pt}
\renewcommand{\arraystretch}{1.15}
\phantomsection\label{ed_table_2}
\noindent \textbf{Extended Data Table 3 | Benchmarks and correctness labels} Stored benchmark splits, sources of correctness labels and the role of each benchmark family in the study.\par\medskip
\begin{longtable}{@{}>{\raggedright\arraybackslash}p{0.1104\linewidth}>{\raggedright\arraybackslash}p{0.2116\linewidth}>{\raggedright\arraybackslash}p{0.2208\linewidth}>{\raggedright\arraybackslash}p{0.3772\linewidth}@{}}
\toprule
Benchmark & Stored source and split & Correctness source & Study role \\
\midrule
\endfirsthead
\toprule
Benchmark & Stored source and split & Correctness source & Study role \\
\midrule
\endhead
MMLU-Pro & Test split; per-row MMLU-Pro source-mixture identifier & Official key and choice parser & Primary modelling set \\
GPQA & Train split; \texttt{Idavidrein/\allowbreak{}gpqa} & Official key and choice parser & Primary modelling set and targeted reranking when correct traces were less frequent \\
MedXpertQA & Test split; \texttt{TsinghuaC3I/\allowbreak{}MedXpertQA} & Official key and choice parser & Primary modelling set \\
MuSR & \texttt{murder\_\allowbreak{}mysteries}, \texttt{object\_\allowbreak{}placements} and \texttt{team\_\allowbreak{}allocation}; \texttt{musr} & Official key and choice parser & Primary modelling set \\
HLE MC & Test split; \texttt{cais/\allowbreak{}hle} & Official key and choice parser & Separate external comparison and targeted reranking when correct traces were less frequent \\
HLE QA & Test split; \texttt{cais/\allowbreak{}hle} & Separate DeepSeek reference-answer equivalence judge & Judge-dependent external diagnostic \\
\bottomrule
\end{longtable}
\endgroup
\section{Data availability}
\label{data-heading}
During peer review, the complete fixed research archive is available to editors and reviewers through the access-controlled Hugging Face dataset repository at \url{https://huggingface.co/datasets/Biogod/mas_reasoning} (revision \texttt{60a16ff6\allowbreak{}026f96d6\allowbreak{}b9da7881\allowbreak{}6d60eeeb\allowbreak{}14d43fdf}). It includes raw model outputs, generated candidate banks, ranked pools, analysis-ready summaries, manifests and the source data underlying all main and Extended Data figures. Reviewer access instructions are supplied separately to the editor. A redistribution-compatible release will be made public upon publication. Any component that cannot be redistributed under the terms of an underlying benchmark or model output will remain available through controlled access.

\section{Code availability}
Analysis, replay and figure-generation code, together with configuration files, tests, provenance records and bilingual reproduction guides, is available at \url{https://github.com/YSTLab/mas-reasoning} (release commit \texttt{7840cf3d\allowbreak{}bde77e61\allowbreak{}a1b4349a\allowbreak{}d3239669\allowbreak{}a7502431}). The repository includes a lock file that identifies the exact Hugging Face data revision used for reproduction. A permanent archived release and identifier will be provided upon publication.

\section{Acknowledgements}
Z.Y. acknowledges the support by National Nature Science Foundation of China (grant numbers 62303119 (Z.Y.) and 32470706 (Z.Y.)), Shanghai Science and Technology Development Funds (grant number 23YF1403000 (Z.Y.)), Fund of Fudan University and Cao'ejiang Basic Research (grant number 24FCA10 (Z.Y.)), the Computational Biology Program (number 25JS2850200 (Z.Y.)) of Science and Technology Commission of Shanghai Municipality (STCSM).

\section{Author contributions}
J.-H.J. conceived the study, designed the experiments, developed the experimental and analysis framework, performed the experiments and statistical analyses, prepared the figures and wrote the manuscript. S.L., J.C. and Z.S. contributed to result validation, interpretation and revision of the manuscript. J.-T.Y. provided biomedical supervision and computational resources. Z.Y. supervised the study, guided the experimental design and overall manuscript structure, reviewed the computational framework and revised the manuscript. All authors reviewed and approved the final manuscript.

\section{Competing interests}
The authors declare no competing interests.

\section{Extended Data}
\label{extended-heading}
{\small\noindent \textbf{a,} Initial comparison of four five-agent communication layouts (\emph{n} = 3,150 trials per layout and stopping condition). The original consensus rule allowed early stopping; a confidence-qualified rule required sufficiently confident agreement; and forced-six-round execution disabled early stopping. Star, complete, ring and hierarchical layouts produced similar accuracy, while token cost rose from about 8-9x with the original rule to 33-49x when all six rounds were required. \textbf{b,} Stopping-round distributions by communication layout. The original early-stopping rule ended 90.48-95.37\% of trials after round 1; even when stopping required confidence-qualified agreement, 81.11-89.17\% ended by round 2. Rounds 4-10 are pooled. Forced-six-round runs are omitted because round 6 was imposed by design. \textbf{c,} Protocol comparison repeated with three random seeds (\emph{n} = 7,350 cases per protocol). The single-answer protocol produced one independently generated response. Majority vote chose the most common of five independently generated answers. In the open committee, five agents answered independently and then answered again after reviewing source-labelled responses from the other four agents. The anonymous evidence board pooled evidence from five independent responses without peer review and removed source labels and explicit links between evidence statements and answer options. Answer-selection details are provided in Methods. Accuracy ranged from 39.88\% for the single-answer protocol to 42.20\% for the open committee, while token cost ranged from 1.00x to 10.73x. \textbf{d,} Answer-selection analysis (\emph{n} = 7,350 per condition). An oracle that counted a case as correct whenever any initial answer was correct exceeded the actual final output by 13.13-14.38 points. When the most common initial answer was wrong but a less frequent correct answer was present, both multi-agent review conditions still selected the most frequent incorrect answer more often than the less frequent correct answer. Source data are provided with this paper.\par}
{\small\noindent \textbf{a,} Benchmark-specific reliability curves for four DeepSeek-V4-Flash (Preview, medium-thinking) and Mimo v2.5 (no-think) combinations. In labels X \ensuremath{\rightarrow} Y, X generated the candidates and Y ranked them; no-think disables Mimo's optional internal reasoning. Rows compare pools with varying correct-wrong counts against pools containing exactly two correct and six wrong candidates (2C/6W). Colour, marker shape and line style identify benchmarks. Rank AUC is the probability that a correct candidate is ranked above a wrong one; 0.5 denotes chance. Solid curves have weighted \emph{R}\textsuperscript{2} \ensuremath{\geq} 0.5. Dashed curves have weighted \emph{R}\textsuperscript{2} < 0.5, so their midpoints are not interpreted. \textbf{b,} Benchmark-specific curves for three Mimo v2.5 (medium-thinking) combinations: DeepSeek generated and Mimo v2.5 ranked; Mimo v2.5 generated and DeepSeek ranked; and Mimo v2.5 both generated and ranked. Medium-thinking enables optional internal reasoning at the medium setting. The two direct no-think/medium-thinking combinations were not run. Comparisons under the same DeepSeek ranking judge were repeated on questions present in both Mimo-generated banks, but each setting still generated its own candidate text. Solid and dashed curves follow the fit-quality convention in a. \textbf{c,} Sensitivity analysis using the same 15,336 questions, 67,163 candidate pools, judge rankings and 823,988 correct-wrong comparisons as the primary DeepSeek-generated and DeepSeek-ranked analysis. Initial-16 availability is the correct fraction among parse-valid traces from the first 16 generation attempts; \emph{p}\textsubscript{gen} remains the full-bank measure. Full-bank and initial-16 availability were strongly correlated (r = 0.994). The respective fitted half-rise midpoints were 14.7\% (95\% CI, 14.3-15.1\%; \emph{R}\textsuperscript{2} = 0.900) and 13.6\% (95\% CI, 13.1-14.0\%; \emph{R}\textsuperscript{2} = 0.927). Shaded bands show 95\% confidence intervals from 1,000 benchmark-stratified question resamples. Source data are provided with this paper; exact curve-level counts are reported in Extended Data Table 1.\par}
{\small\noindent \textbf{a,} Humanity's Last Exam multiple-choice (HLE MC) reliability curves for four DeepSeek-V4-Flash (Preview, medium-thinking)/Mimo v2.5 (no-think) generator and ranking-judge combinations, using official answer keys. In labels X -> Y, X generated the candidates and Y ranked them. Rank AUC is the pair-weighted probability that a correct candidate is ranked above a wrong one; 0.5 denotes chance. Solid curves have \emph{R}\textsuperscript{2} \ensuremath{\geq} 0.5; dashed curves have lower fit. \textbf{b,} Humanity's Last Exam multiple-choice (HLE MC) reliability curves for the three observed combinations involving Mimo v2.5 (medium-thinking). Medium-thinking enables optional internal reasoning at the medium setting. The two direct no-think/medium-thinking combinations were not run. Official answer keys determine correctness; points and fit styles follow a. HLE was excluded from the four-benchmark fit and tests whether the reliability relationship depends on the benchmark and model setting. \textbf{c,} Humanity's Last Exam open-answer (HLE QA) reliability curves for four DeepSeek-V4-Flash (Preview, medium-thinking)/Mimo v2.5 (no-think) generator and ranking-judge combinations. A separate model compared each generated answer with the reference to assign correctness; the ranking judge saw neither the reference nor those labels. Candidate banks contained 143,977 valid DeepSeek and 164,176 valid Mimo answer-and-rationale candidates across 1,645 questions. An author-reviewed audit included 127 generated candidates and 200 controls. After three ambiguous generated cases were excluded, agreement was 96.8\% (120/124) among the remaining generated cases and 97.5\% (195/200) across the controls. The curves reflect model-based equivalence labels assigned to open-answer candidates. Source data are provided with this paper.\par}
{\small\noindent \textbf{a,} Coverage of the offline comparison that reused stored candidates and judge orderings without new model calls. DeepSeek-V4-Flash (Preview, medium-thinking) generated and ranked five pools of eight candidates (\emph{k} = 8) per question, yielding 16,278 questions and 81,390 pools. The judge returned a usable complete ordering for every retained pool; no question was lost because of a missing or invalid ranking response. \textbf{b,} Question-level accuracy for one candidate sampled from the full bank, a control that randomly shuffles the judge's order, majority vote over the stored pool and selection of the judge's first-ranked candidate. Results are shown across five benchmarks using the same offline candidate pools. Bars are point estimates; paired uncertainty is shown in c. \textbf{c,} Change in accuracy when first-ranked selection replaced majority vote on the same candidate pools. Points are mean paired question-level differences by benchmark and bars are conditional 95\% CIs from 1,000 question bootstraps. All five estimates are positive. \textbf{d,} Gain over one candidate sampled from the full bank across nine bins of \emph{p}\textsubscript{gen}, the question-level correct fraction in that bank. Curves show majority vote, the shuffled-order control and first-ranked selection. Values are descriptive binned point estimates rather than fitted thresholds or universal availability ranges. Source data are provided with this paper.\par}
{\small\noindent \textbf{a,} Matched DeepSeek-V4-Flash (Preview, medium-thinking) reranking of 713 pools, each containing at least one correct candidate but fewer correct than wrong candidates. This does not mean that one wrong answer receives most votes. The same questions, candidate labels and displayed text are retained across five conditions: answer plus rationale, answer only, rationale only, answers reassigned among candidate labels or rationales reassigned among labels. For example, an answer originally attached to candidate A may be shown under candidate B. Reassignment preserves all text and formatting but breaks its original candidate link. \textbf{b,} The probability that a correct candidate was ranked above a wrong one, termed rank AUC, was measured under five information conditions (\emph{n} = 713 matched pools; 6,488 correct-wrong comparisons per condition; 100\% usable rankings); 0.5 denotes chance. Rank AUC fell from 0.654 when answers remained with their original candidates to 0.568 when answers were reassigned, and from 0.655 to 0.482 when supporting rationales were reassigned. \textbf{c,} Change in the probability that a correct candidate ranked above a wrong one when content remained with its original candidate rather than being reassigned. Across the same 713 pools, rank-AUC differences were +0.173 for supporting rationales and +0.086 for answers. Confidence intervals used 2,000 paired pool-level bootstrap resamples. Intervals included zero among questions with the smallest \emph{p}\textsubscript{gen} values, where \emph{p}\textsubscript{gen} is the correct fraction in the full candidate bank. The other samples target questions nearest the fitted half-rise or with both candidate diversity and above-chance fitted signal; these are diagnostic question-selection rules, not universal bins. Source data are provided with this paper.\par}
{\small\noindent \textbf{a,} Accuracy-cost curves for pools generated by Mimo v2.5 with optional internal reasoning disabled (no-think) or enabled at medium (medium-thinking). At each pool size, each comparison retains questions with valid outputs for every displayed answer-selection method. DS denotes DeepSeek-V4-Flash (Preview, medium-thinking). In slash labels, the generator comes first and the ranking judge second; Mimo/DS therefore means that Mimo generated the candidates and DeepSeek ranked them. Curves compare first-ranked selection with majority vote on the same pools. Costs assume that 90\% of repeated input receives the lower cached-input price and 10\% receives the full input price. Panel d also compares \emph{k} = 16 on exactly the same 11,241 questions. \textbf{b,} Change in accuracy when first-ranked selection replaced majority vote on the same pools across \emph{k}. At \emph{k} = 16, restricting both generator settings to the same 11,241 questions yielded +37.44 points for Mimo no-think pools and +2.96 points for Mimo medium-thinking pools when DeepSeek ranked them. The questions were matched, but each setting generated its own candidate answers and rationales. \textbf{c,} Vote structure at \emph{k} = 16: no correct answer; an incorrect answer gets the highest count; the correct answer ties for the highest count; or only the correct answer gets the highest count. The highest count need not exceed 50\% of the pool. DeepSeek-generated and Mimo no-think-generated bars contain 11,241 pools from the shared comparison set; the Mimo v2.5 medium-thinking-generated bar contains its full valid-output set of 16,247 pools. The panel distinguishes the presence of a correct answer from majority vote's ability to select it. \textbf{d,} Majority vote, DeepSeek first-ranked selection and oracle accuracy at \emph{k} = 16 on the same 11,241 questions for both Mimo generator settings. DeepSeek ranking exceeded majority vote by 37.44 points for no-think generation and 2.96 points for medium-thinking generation. The questions were matched, but each setting generated its own answers and rationales. \textbf{e,} Each preserved natural \emph{k} = 16 DeepSeek candidate pool was truncated to every pool size from \emph{k} = 2 to 16 across the same 11,241 questions. Each point compares an odd pool size with the midpoint of its adjacent even pool sizes on the same questions. Odd-pool majority-accuracy deviations remained within 0.39 percentage points of zero, and all seven paired 95\% confidence intervals included zero, whereas odd pools reduced plurality-tie rates by 1.64-5.96 points. The offline diagnostic reused stored candidates and made no new model calls. Source data are provided with this paper.\par}
\begin{figure}[H]
\centering
\phantomsection\label{ed_fig7}
\includegraphics[alt={Extended Data Fig. 7 | Useful judge influence varies across benchmarks and model roles},width=\linewidth]{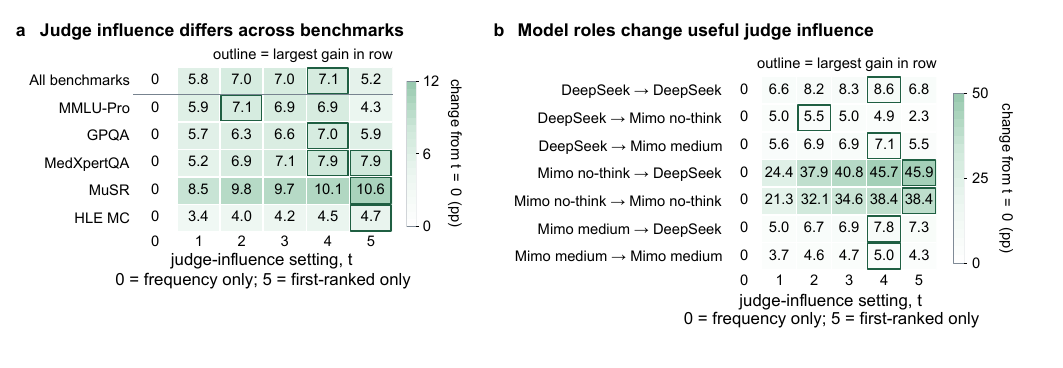}
\caption*{\textbf{Extended Data Fig. 7 | Useful judge influence varies across benchmarks and model roles}}
\end{figure}
{\small\noindent \textbf{a,} Accuracy change relative to majority vote across judge-influence settings in the primary analysis (16,278 questions; five stored pools of eight candidates per question). Setting t = 0 uses answer frequency alone, t = 1-4 progressively increases the contribution of the judge's ranking while retaining frequency information, and t = 5 selects the answer attached to the judge's first-ranked candidate. The pooled row is separated from the five benchmark rows. An outline marks each row's largest observed gain; this panel uses its own colour scale. The setting with the highest observed accuracy differed across benchmarks. \textbf{b,} Global accuracy change relative to t = 0 across seven generator and ranking-judge combinations. The model before the arrow generated candidates and the model after it ranked them. This analysis includes only pools containing at least one correct candidate and a complete judge ranking. An outline marks each row's largest observed gain; this panel uses its own colour scale, which differs from a. Stronger judge influence produced the largest gains for Mimo no-think-generated pools, whereas moderate settings often performed best when DeepSeek generated the candidates. The highest observed setting varied across model roles. Source data are provided with this paper.\par}
\begin{landscape}
\section{Extended Data Tables}
\begingroup
\fontsize{5.2}{6.1}\selectfont
\setlength{\tabcolsep}{1.4pt}
\renewcommand{\arraystretch}{1.15}
\phantomsection\label{ed_table1}
\noindent \textbf{Extended Data Table 1 | Benchmark- and model-specific judge-reliability fits} Fit parameters, sample sizes and interpretation status for each observed benchmark, pool-construction setting and generator-ranking-model combination. Midpoints are interpreted only when the weighted fit has \emph{R}\textsuperscript{2} \ensuremath{\geq} 0.5.\par\medskip
\begin{longtable}{@{}>{\raggedright\arraybackslash}p{0.0664\linewidth}>{\raggedright\arraybackslash}p{0.0872\linewidth}>{\raggedright\arraybackslash}p{0.0830\linewidth}>{\raggedright\arraybackslash}p{0.0830\linewidth}>{\raggedright\arraybackslash}p{0.0374\linewidth}>{\raggedright\arraybackslash}p{0.0415\linewidth}>{\raggedright\arraybackslash}p{0.0664\linewidth}>{\raggedright\arraybackslash}p{0.0498\linewidth}>{\raggedright\arraybackslash}p{0.0539\linewidth}>{\raggedright\arraybackslash}p{0.0457\linewidth}>{\raggedright\arraybackslash}p{0.0290\linewidth}>{\raggedright\arraybackslash}p{0.0457\linewidth}>{\raggedright\arraybackslash}p{0.0539\linewidth}>{\raggedright\arraybackslash}p{0.0872\linewidth}@{}}
\toprule
Dataset & Pool construction & Generator & Judge & AUC & Fit \emph{R}\textsuperscript{2} & Fit quality & Midpoint & Pairs & Pools & Bins & Parse failures & Fit status & Evidence use \\
\midrule
\endfirsthead
\toprule
Dataset & Pool construction & Generator & Judge & AUC & Fit \emph{R}\textsuperscript{2} & Fit quality & Midpoint & Pairs & Pools & Bins & Parse failures & Fit status & Evidence use \\
\midrule
\endhead
MMLU-Pro & Sampled composition & DeepSeek-V4-Flash (Preview, medium-thinking) & DeepSeek-V4-Flash (Preview, medium-thinking) & 0.907 & 0.944 & strong boundary fit & 13.8\% & 645,441 & 52,097 & 31 & 0 & ok & strong benchmark-specific formula fit \\
MMLU-Pro & Sampled composition & DeepSeek-V4-Flash (Preview, medium-thinking) & Mimo v2.5 (no-think) & 0.886 & 0.939 & strong boundary fit & 13.4\% & 645,429 & 52,096 & 31 & 1 & ok & strong benchmark-specific formula fit \\
MMLU-Pro & Sampled composition & Mimo v2.5 (no-think) & DeepSeek-V4-Flash (Preview, medium-thinking) & 0.896 & 0.127 & low-\emph{R}\textsuperscript{2} diagnostic & 0.5\% & 383,093 & 43,214 & 20 & 0 & ok & AUC >= 0.85, low-\emph{R}\textsuperscript{2} diagnostic \\
MMLU-Pro & Sampled composition & Mimo v2.5 (no-think) & Mimo v2.5 (no-think) & 0.880 & 0.257 & low-\emph{R}\textsuperscript{2} diagnostic & 1.6\% & 383,074 & 43,212 & 20 & 2 & ok & AUC >= 0.85, low-\emph{R}\textsuperscript{2} diagnostic \\
MMLU-Pro & Sampled composition & DeepSeek-V4-Flash (Preview, medium-thinking) & Mimo v2.5 (medium-thinking) & 0.901 & 0.955 & strong boundary fit & 14.8\% & 645,413 & 52,095 & 31 & 2 & ok & strong benchmark-specific formula fit \\
MMLU-Pro & Sampled composition & Mimo v2.5 (medium-thinking) & DeepSeek-V4-Flash (Preview, medium-thinking) & 0.884 & 0.942 & strong boundary fit & 16.8\% & 414,911 & 36,925 & 49 & 1 & ok & strong benchmark-specific formula fit \\
MMLU-Pro & Sampled composition & Mimo v2.5 (medium-thinking) & Mimo v2.5 (medium-thinking) & 0.869 & 0.955 & strong boundary fit & 20.8\% & 414,916 & 36,925 & 49 & 1 & ok & strong benchmark-specific formula fit \\
MMLU-Pro & Fixed composition & DeepSeek-V4-Flash (Preview, medium-thinking) & DeepSeek-V4-Flash (Preview, medium-thinking) & 0.858 & 0.919 & strong boundary fit & 12.6\% & 106,608 & 8,884 & 26 & 0 & ok & strong benchmark-specific formula fit \\
MMLU-Pro & Fixed composition & DeepSeek-V4-Flash (Preview, medium-thinking) & Mimo v2.5 (no-think) & 0.841 & 0.933 & strong boundary fit & 13.1\% & 106,596 & 8,883 & 26 & 1 & ok & strong benchmark-specific formula fit \\
MMLU-Pro & Fixed composition & Mimo v2.5 (no-think) & DeepSeek-V4-Flash (Preview, medium-thinking) & 0.902 & 0.001 & low-\emph{R}\textsuperscript{2} diagnostic & 1.3\% & 96,672 & 8,056 & 18 & 0 & ok & AUC >= 0.85, low-\emph{R}\textsuperscript{2} diagnostic \\
MMLU-Pro & Fixed composition & Mimo v2.5 (no-think) & Mimo v2.5 (no-think) & 0.891 & 0.051 & low-\emph{R}\textsuperscript{2} diagnostic & 3.8\% & 96,672 & 8,056 & 18 & 0 & ok & AUC >= 0.85, low-\emph{R}\textsuperscript{2} diagnostic \\
MMLU-Pro & Fixed composition & DeepSeek-V4-Flash (Preview, medium-thinking) & Mimo v2.5 (medium-thinking) & 0.846 & 0.917 & strong boundary fit & 13.6\% & 106,608 & 8,884 & 26 & 0 & ok & strong benchmark-specific formula fit \\
MMLU-Pro & Fixed composition & Mimo v2.5 (medium-thinking) & DeepSeek-V4-Flash (Preview, medium-thinking) & 0.848 & 0.888 & strong boundary fit & 15.6\% & 89,472 & 7,456 & 45 & 1 & ok & strong benchmark-specific formula fit \\
MMLU-Pro & Fixed composition & Mimo v2.5 (medium-thinking) & Mimo v2.5 (medium-thinking) & 0.824 & 0.907 & strong boundary fit & 20.0\% & 89,484 & 7,457 & 45 & 0 & ok & strong benchmark-specific formula fit \\
GPQA & Sampled composition & DeepSeek-V4-Flash (Preview, medium-thinking) & DeepSeek-V4-Flash (Preview, medium-thinking) & 0.836 & 0.860 & strong boundary fit & 29.8\% & 29,796 & 2,337 & 24 & 0 & ok & strong benchmark-specific formula fit \\
GPQA & Sampled composition & DeepSeek-V4-Flash (Preview, medium-thinking) & Mimo v2.5 (no-think) & 0.745 & 0.848 & strong boundary fit & 31.2\% & 29,796 & 2,337 & 24 & 0 & ok & strong benchmark-specific formula fit \\
GPQA & Sampled composition & Mimo v2.5 (no-think) & DeepSeek-V4-Flash (Preview, medium-thinking) & 0.791 & 0.687 & usable boundary fit & 20.7\% & 30,544 & 2,488 & 19 & 0 & ok & usable benchmark-specific transition \\
GPQA & Sampled composition & Mimo v2.5 (no-think) & Mimo v2.5 (no-think) & 0.728 & 0.700 & usable boundary fit & 23.2\% & 30,544 & 2,488 & 19 & 0 & ok & usable benchmark-specific transition \\
GPQA & Sampled composition & DeepSeek-V4-Flash (Preview, medium-thinking) & Mimo v2.5 (medium-thinking) & 0.815 & 0.855 & strong boundary fit & 32.1\% & 29,796 & 2,337 & 24 & 0 & ok & strong benchmark-specific formula fit \\
GPQA & Sampled composition & Mimo v2.5 (medium-thinking) & DeepSeek-V4-Flash (Preview, medium-thinking) & 0.785 & 0.855 & strong boundary fit & 29.5\% & 24,374 & 2,036 & 35 & 4 & ok & strong benchmark-specific formula fit \\
GPQA & Sampled composition & Mimo v2.5 (medium-thinking) & Mimo v2.5 (medium-thinking) & 0.762 & 0.835 & strong boundary fit & 34.6\% & 24,424 & 2,040 & 35 & 0 & ok & strong benchmark-specific formula fit \\
GPQA & Fixed composition & DeepSeek-V4-Flash (Preview, medium-thinking) & DeepSeek-V4-Flash (Preview, medium-thinking) & 0.725 & 0.736 & usable boundary fit & 28.9\% & 4,224 & 352 & 18 & 0 & ok & usable benchmark-specific transition \\
GPQA & Fixed composition & DeepSeek-V4-Flash (Preview, medium-thinking) & Mimo v2.5 (no-think) & 0.619 & 0.610 & usable boundary fit & 35.1\% & 4,224 & 352 & 18 & 0 & ok & usable benchmark-specific transition \\
GPQA & Fixed composition & Mimo v2.5 (no-think) & DeepSeek-V4-Flash (Preview, medium-thinking) & 0.781 & 0.888 & strong boundary fit & 22.9\% & 5,760 & 480 & 12 & 0 & ok & strong benchmark-specific formula fit \\
GPQA & Fixed composition & Mimo v2.5 (no-think) & Mimo v2.5 (no-think) & 0.710 & 0.647 & usable boundary fit & 34.6\% & 5,760 & 480 & 12 & 0 & ok & usable benchmark-specific transition \\
GPQA & Fixed composition & DeepSeek-V4-Flash (Preview, medium-thinking) & Mimo v2.5 (medium-thinking) & 0.690 & 0.669 & usable boundary fit & 34.9\% & 4,224 & 352 & 18 & 0 & ok & usable benchmark-specific transition \\
GPQA & Fixed composition & Mimo v2.5 (medium-thinking) & DeepSeek-V4-Flash (Preview, medium-thinking) & 0.734 & 0.761 & strong boundary fit & 31.1\% & 5,076 & 423 & 32 & 1 & ok & strong benchmark-specific formula fit \\
GPQA & Fixed composition & Mimo v2.5 (medium-thinking) & Mimo v2.5 (medium-thinking) & 0.699 & 0.758 & strong boundary fit & 34.3\% & 5,088 & 424 & 32 & 0 & ok & strong benchmark-specific formula fit \\
MedXpertQA & Sampled composition & DeepSeek-V4-Flash (Preview, medium-thinking) & DeepSeek-V4-Flash (Preview, medium-thinking) & 0.779 & 0.921 & strong boundary fit & 15.0\% & 98,089 & 9,065 & 28 & 0 & ok & strong benchmark-specific formula fit \\
MedXpertQA & Sampled composition & DeepSeek-V4-Flash (Preview, medium-thinking) & Mimo v2.5 (no-think) & 0.703 & 0.911 & strong boundary fit & 18.1\% & 98,089 & 9,065 & 28 & 0 & ok & strong benchmark-specific formula fit \\
MedXpertQA & Sampled composition & Mimo v2.5 (no-think) & DeepSeek-V4-Flash (Preview, medium-thinking) & 0.763 & 0.778 & strong boundary fit & 12.4\% & 66,433 & 8,165 & 7 & 0 & ok & strong benchmark-specific formula fit \\
MedXpertQA & Sampled composition & Mimo v2.5 (no-think) & Mimo v2.5 (no-think) & 0.702 & 0.812 & strong boundary fit & 17.0\% & 66,433 & 8,165 & 7 & 0 & ok & strong benchmark-specific formula fit \\
MedXpertQA & Sampled composition & DeepSeek-V4-Flash (Preview, medium-thinking) & Mimo v2.5 (medium-thinking) & 0.746 & 0.930 & strong boundary fit & 18.5\% & 98,089 & 9,065 & 28 & 0 & ok & strong benchmark-specific formula fit \\
MedXpertQA & Sampled composition & Mimo v2.5 (medium-thinking) & DeepSeek-V4-Flash (Preview, medium-thinking) & 0.752 & 0.926 & strong boundary fit & 15.5\% & 88,646 & 8,841 & 43 & 0 & ok & strong benchmark-specific formula fit \\
MedXpertQA & Sampled composition & Mimo v2.5 (medium-thinking) & Mimo v2.5 (medium-thinking) & 0.706 & 0.923 & strong boundary fit & 18.0\% & 88,646 & 8,841 & 43 & 0 & ok & strong benchmark-specific formula fit \\
MedXpertQA & Fixed composition & DeepSeek-V4-Flash (Preview, medium-thinking) & DeepSeek-V4-Flash (Preview, medium-thinking) & 0.698 & 0.896 & strong boundary fit & 12.5\% & 27,636 & 2,303 & 23 & 0 & ok & strong benchmark-specific formula fit \\
MedXpertQA & Fixed composition & DeepSeek-V4-Flash (Preview, medium-thinking) & Mimo v2.5 (no-think) & 0.641 & 0.820 & strong boundary fit & 12.3\% & 27,636 & 2,303 & 23 & 0 & ok & strong benchmark-specific formula fit \\
MedXpertQA & Fixed composition & Mimo v2.5 (no-think) & DeepSeek-V4-Flash (Preview, medium-thinking) & 0.791 & 0.769 & strong boundary fit & 12.6\% & 17,724 & 1,477 & 5 & 0 & ok & strong benchmark-specific formula fit \\
MedXpertQA & Fixed composition & Mimo v2.5 (no-think) & Mimo v2.5 (no-think) & 0.745 & 0.882 & strong boundary fit & 14.3\% & 17,724 & 1,477 & 5 & 0 & ok & strong benchmark-specific formula fit \\
MedXpertQA & Fixed composition & DeepSeek-V4-Flash (Preview, medium-thinking) & Mimo v2.5 (medium-thinking) & 0.670 & 0.878 & strong boundary fit & 13.4\% & 27,636 & 2,303 & 23 & 0 & ok & strong benchmark-specific formula fit \\
MedXpertQA & Fixed composition & Mimo v2.5 (medium-thinking) & DeepSeek-V4-Flash (Preview, medium-thinking) & 0.710 & 0.830 & strong boundary fit & 13.8\% & 27,552 & 2,296 & 40 & 0 & ok & strong benchmark-specific formula fit \\
MedXpertQA & Fixed composition & Mimo v2.5 (medium-thinking) & Mimo v2.5 (medium-thinking) & 0.670 & 0.847 & strong boundary fit & 15.3\% & 27,552 & 2,296 & 40 & 0 & ok & strong benchmark-specific formula fit \\
MuSR & Sampled composition & DeepSeek-V4-Flash (Preview, medium-thinking) & DeepSeek-V4-Flash (Preview, medium-thinking) & 0.734 & 0.422 & low-\emph{R}\textsuperscript{2} diagnostic & 56.4\% & 50,662 & 3,664 & 19 & 0 & ok & intermediate-AUC, low-\emph{R}\textsuperscript{2} diagnostic \\
MuSR & Sampled composition & DeepSeek-V4-Flash (Preview, medium-thinking) & Mimo v2.5 (no-think) & 0.656 & 0.686 & usable boundary fit & 53.5\% & 50,662 & 3,664 & 19 & 0 & ok & usable benchmark-specific transition \\
MuSR & Sampled composition & Mimo v2.5 (no-think) & DeepSeek-V4-Flash (Preview, medium-thinking) & 0.736 & 0.006 & low-\emph{R}\textsuperscript{2} diagnostic & 8.4\% & 50,392 & 3,721 & 15 & 0 & ok & intermediate-AUC, low-\emph{R}\textsuperscript{2} diagnostic \\
MuSR & Sampled composition & Mimo v2.5 (no-think) & Mimo v2.5 (no-think) & 0.680 & 0.006 & low-\emph{R}\textsuperscript{2} diagnostic & 6.1\% & 50,392 & 3,721 & 15 & 0 & ok & intermediate-AUC, low-\emph{R}\textsuperscript{2} diagnostic \\
MuSR & Sampled composition & DeepSeek-V4-Flash (Preview, medium-thinking) & Mimo v2.5 (medium-thinking) & 0.707 & 0.592 & usable boundary fit & 59.0\% & 50,662 & 3,664 & 19 & 0 & ok & usable benchmark-specific transition \\
MuSR & Sampled composition & Mimo v2.5 (medium-thinking) & DeepSeek-V4-Flash (Preview, medium-thinking) & 0.775 & 0.679 & usable boundary fit & 45.7\% & 51,695 & 3,706 & 30 & 0 & ok & usable benchmark-specific transition \\
MuSR & Sampled composition & Mimo v2.5 (medium-thinking) & Mimo v2.5 (medium-thinking) & 0.737 & 0.718 & usable boundary fit & 49.7\% & 51,695 & 3,706 & 30 & 0 & ok & usable benchmark-specific transition \\
MuSR & Fixed composition & DeepSeek-V4-Flash (Preview, medium-thinking) & DeepSeek-V4-Flash (Preview, medium-thinking) & 0.557 & 0.214 & low-\emph{R}\textsuperscript{2} diagnostic & 54.6\% & 7,740 & 645 & 14 & 0 & ok & AUC <= 0.65, low-\emph{R}\textsuperscript{2} diagnostic \\
MuSR & Fixed composition & DeepSeek-V4-Flash (Preview, medium-thinking) & Mimo v2.5 (no-think) & 0.524 & 0.266 & low-\emph{R}\textsuperscript{2} diagnostic & 57.5\% & 7,740 & 645 & 14 & 0 & ok & AUC <= 0.65, low-\emph{R}\textsuperscript{2} diagnostic \\
MuSR & Fixed composition & Mimo v2.5 (no-think) & DeepSeek-V4-Flash (Preview, medium-thinking) & 0.684 & 0.000 & low-\emph{R}\textsuperscript{2} diagnostic & 0.1\% & 9,060 & 755 & 14 & 0 & ok & intermediate-AUC, low-\emph{R}\textsuperscript{2} diagnostic \\
MuSR & Fixed composition & Mimo v2.5 (no-think) & Mimo v2.5 (no-think) & 0.625 & 0.000 & low-\emph{R}\textsuperscript{2} diagnostic & 0.1\% & 9,060 & 755 & 14 & 0 & ok & AUC <= 0.65, low-\emph{R}\textsuperscript{2} diagnostic \\
MuSR & Fixed composition & DeepSeek-V4-Flash (Preview, medium-thinking) & Mimo v2.5 (medium-thinking) & 0.541 & 0.084 & low-\emph{R}\textsuperscript{2} diagnostic & 22.8\% & 7,740 & 645 & 14 & 0 & ok & AUC <= 0.65, low-\emph{R}\textsuperscript{2} diagnostic \\
MuSR & Fixed composition & Mimo v2.5 (medium-thinking) & DeepSeek-V4-Flash (Preview, medium-thinking) & 0.687 & 0.552 & usable boundary fit & 58.0\% & 9,048 & 754 & 29 & 0 & ok & usable benchmark-specific transition \\
MuSR & Fixed composition & Mimo v2.5 (medium-thinking) & Mimo v2.5 (medium-thinking) & 0.615 & 0.701 & usable boundary fit & 57.2\% & 9,048 & 754 & 29 & 0 & ok & usable benchmark-specific transition \\
\bottomrule
\end{longtable}
\endgroup
\end{landscape}
\end{document}